\documentclass{article}

\PassOptionsToPackage{numbers, compress}{natbib}

\usepackage[preprint]{neurips_2026}

\usepackage[utf8]{inputenc} 
\usepackage[T1]{fontenc}    
\usepackage{hyperref}       
\usepackage{url}            
\usepackage{booktabs}       
\usepackage{amsfonts}       
\usepackage{nicefrac}       
\usepackage{microtype}      
\usepackage{xcolor}         
\usepackage{algorithm}
\usepackage{algpseudocode}
\usepackage{wrapfig}
\usepackage{capt-of}
\usepackage{float}

\usepackage{graphicx}
\usepackage{amsmath}
\usepackage{utfsym}  
\usepackage{multirow,algorithm}
\usepackage{colortbl}
\definecolor{linecolor2}{RGB}{230,245,245}
\usepackage{enumitem}
\usepackage{amssymb}
\usepackage{makecell}
\usepackage{wrapfig}
\usepackage{bm}
\usepackage{tikz}
\usepackage{threeparttablex}
\usepackage{tabularx}

\usepackage{titletoc}

\title{
ReWorld: Multi-Dimensional Reward Modeling for Embodied Video Generation
}

\author{%
  \begin{minipage}{0.96\textwidth}
  \centering
  \normalsize
  Baorui~Peng$^{1,2}$\thanks{Equal contribution.}\quad
  Wenyao~Zhang$^{1,3}$\footnotemark[1]\thanks{Project leader.}\quad
  Liang~Xu$^{1,3}$\quad
  Bohan~Li$^{1,3}$\\[0.3em]
  Hu~Zhu$^{1,7}$\quad
  Zekun~Qi$^{4}$\quad
  Hongsi~Liu$^{1,5}$\quad
  Jiazhao~Zhang$^{6}$\\[0.3em]
  Wenjun~Zeng$^{1}$\quad
  Xin~Jin$^{1}$\thanks{Corresponding author.}\\[0.7em]
  \small\normalfont
  $^{1}$Eastern Institute of Technology, Ningbo\\
  $^{2}$Georgia Institute of Technology\quad
  $^{3}$Shanghai Jiao Tong University\\
  $^{4}$Tsinghua University\quad
  $^{5}$University of Science and Technology of China\\
  $^{6}$Peking University\quad
  $^{7}$The Hong Kong Polytechnic University
  \end{minipage}%
}

\begin{document}

\maketitle

\begin{abstract}
 \vspace{-10pt}
Current embodied video generation models primarily emphasize visual generative quality while overlooking physical fidelity, dynamic consistency, and task logic, especially for contact-rich manipulation tasks, which limits their applicability to downstream tasks.
In this paper, we introduce ReWorld, a framework aimed to employ reinforcement learning to align the video generation with \textbf{physical realism, task completion capability, embodiment plausibility} and \textbf{visual quality.}
Specifically, we first construct a large-scale ($\sim$$235K$) video preference dataset and employ it to train a hierarchical reward model designed to capture multi-dimensional reward consistent with human preferences.
We further propose a practical alignment algorithm that post-trains flow-based video generators using this reward through a computationally efficient PPO-style algorithm.
Comprehensive experiments and theoretical analysis demonstrate that ReWorld significantly improves the physical fidelity, logical coherence, embodiment and visual quality of generated rollouts, outperforming previous methods.\looseness=-1
\end{abstract}

\section{Introduction}

Embodied video generation models~\cite{ha2018world, hafner2023mastering} that learn to simulate the dynamics of embodied environments, have become a cornerstone for developing general-purpose embodied intelligence~\cite{liao2025genie,chi2025wow}. 
They enable diverse downstream capabilities, including scalable data generation~\cite{li2025uniscene}, interactive simulation~\cite{chandra2025diwa,li2025vla}, and integration with vision-language-action models~\cite{vpp2024hu,zhang2025dreamvla}.

\begin{figure}[t]
  \centering
  \includegraphics[width=0.86\linewidth]{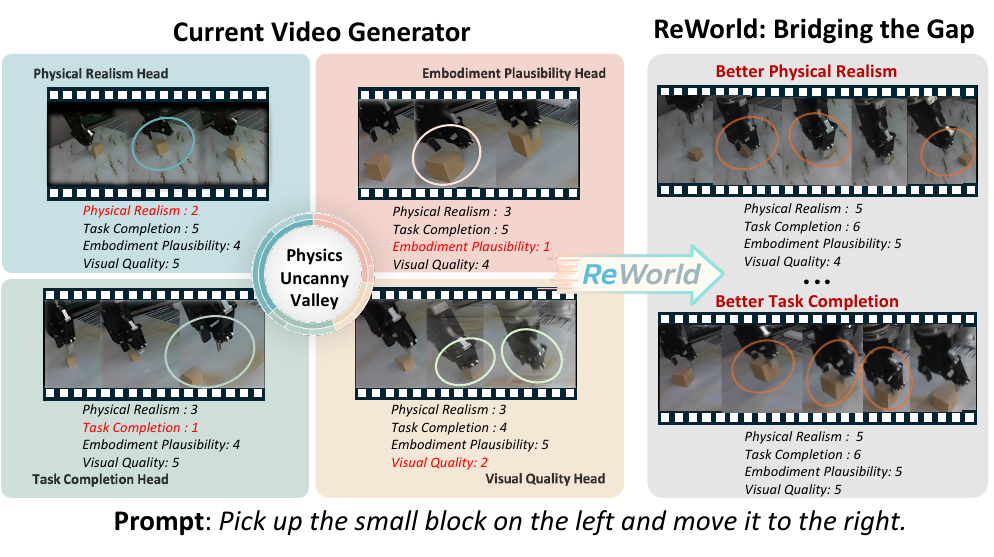}
  \vspace{-3mm}
   \caption{ReWorld bridges the gap of embodied video generator with \textbf{physical realism, task completion capability, embodiment plausibility,} and \textbf{visual quality.} Each of the four dimensions shown is rated on a scale of 1 (poor) to 6 (excellent), where higher scores indicate better performance.}
   \vspace{-1mm}
   \label{fig:teaser}
\end{figure}

Despite the impressive generative capabilities~\cite{alhaija2025cosmos,sora2024}, they still struggle with what we term as \textbf{Physics Uncanny Valley}—a gap between visual plausibility and physical consistency as shown in~\ref{fig:teaser}. 
This limitation primarily arises because current models are trained almost exclusively under a supervised paradigm, exposing them only to successful demonstrations. 
Without encountering failures or understanding ``what not to do'', these models struggle to internalize the implicit physical laws that govern the real world.
Although previous works try to introduce more abundant conditions (\textit{e.g.}, text~\cite{wang2023videocomposer,chen2023control}, trajectory~\cite{fu2025learning}, depth~\cite{yang2025orv,zhen2025tesseract} and physical law~\cite{liu2024physgen}), maintaining visual plausibility and physical consistency remains a formidable challenge.

This raises a key question: how can we align these powerful generative models with the complex, implicit rules of physical interaction?
Reinforcement learning (RL)~\cite{sutton1998reinforcement}, particularly when combined with human feedback (RLHF)~\cite{christiano2017deep}, has successfully addressed similar alignment challenges in other domains(\textit{e.g.}, Large Language Models~\cite{wang2023videocomposer,chen2023control}, image generation~\cite{brooks2023instructpix2pix,ye2024dreamreward}). 
However, transferring this paradigm to robotics videos suffers from two major barriers:
(i)~\textbf{The Reward Barrier (Perception):} How do we define a ``good'' embodied video? 
Sparse task-success signals from online interaction are insufficient~\cite{chi2025wow}, as they fail to penalize subtle yet critical physical violations. 
Conversely, a single, monolithic reward like aesthetic score~\cite{ye2024dreamreward} or CLIP score~\cite{radford2021learning} is also inadequate. 
It cannot simultaneously evaluate low-level physics like \textit{``did the hand penetrate the cup?''} and high-level semantics like \textit{``did the agent pick up the correct cup?''}. 
(ii)~\textbf{The Algorithm Barrier (Optimization):} 
Even with a perfect reward, how can we effectively optimize a flow-based video generation model--the dominant paradigm in current video generation research?
Previous methods often focus on simpler, task-specific rewards, rather than the complex, multi-dimensional requirements of robotics~\cite{hafner2023mastering,xue2025dancegrpo}. 
Refining flow-based models~\cite{alhaija2025cosmos} remains a critical yet unsolved problem. 
This is because standard policy gradient methods like PPO~\cite{schulman2017proximal} rely on computing the log-likelihood term $\log \pi_\theta(v|c)$. 
However, for flow-based models, this computation is prohibitively expensive, as it involves an $\mathcal{O}(d^2 \cdot T_{\text{ODE}})$ integration over the Jacobian trace~\cite{mcallister2025flow}.
This computational bottleneck makes PPO-style optimization practically infeasible for flow matching-based models~\cite{alhaija2025cosmos}.

\begin{figure*}[t]
  \centering
   \vspace{2mm}
   \includegraphics[width=1.0\linewidth]{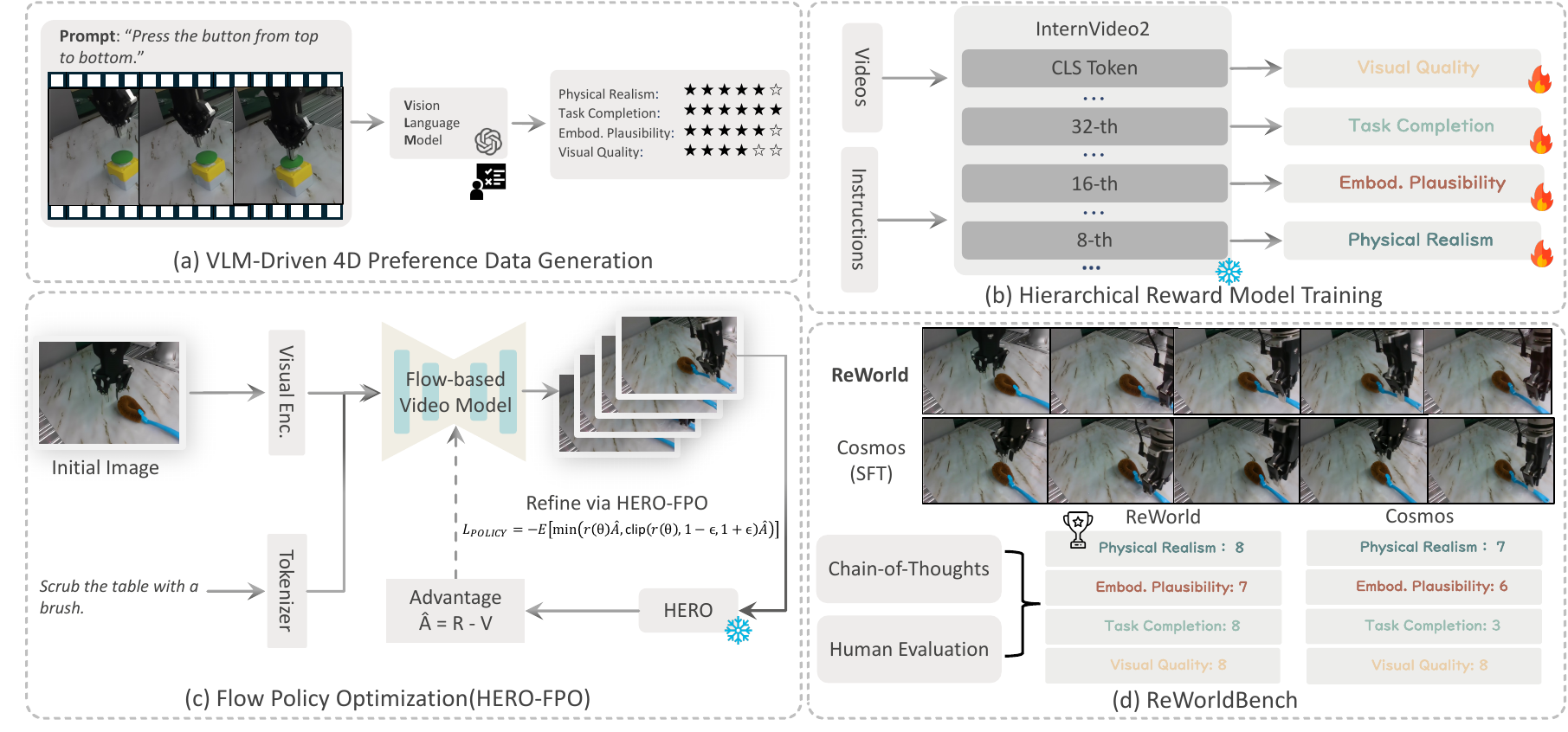}
   \vspace{-5mm}
   \caption{\textbf{Overview of the ReWorld framework}. (a) We employ a VLM-driven annotation system to generate the 4-dimensional embodied preference dataset. (b) Building upon this dataset, we train the multi-dimensional reward model HERO based on the hierarchical feature space of InternVideo2~\cite{wang2024internvideo2}. (c) We detail the reinforcement learning pipeline HERO-FPO to refine the generative policy with the learned multi-dimensional reward signal. (d) We introduce ReWorldBench as a specialized benchmark to evaluate embodied video generation models.}
   \label{fig:hero_architecture}
   \vspace{-1mm}
\end{figure*}

To this end, we propose ReWorld, a new framework that systematically bridges both barriers through a multi-dimensional reward model and a flow-based video generator optimization approach to align embodied video generation models with implicit physical realism.
As shown in Fig.~\ref{fig:teaser}, we first construct a large-scale video preference dataset to capture human preference over physical realism, embodiment plausibility, and task semantics.
Building upon this foundation, we introduce HERO (HiErarchical Reward mOdel), a multi-dimensional reward model.
Its core innovation is multi-dimensional reward awareness: a decoupled, four-head architecture to specialize in physical fidelity, embodiment plausibility, task completion, and visual quality, respectively. 
While VLMs can provide useful supervision, using an off-the-shelf VLM directly as an online reward is costly and can be unstable; HERO distills VLM-based supervision into a fast, video-native reward model suitable for RL fine-tuning.
Critically, as shown in Fig.~\ref{fig:hero_architecture}, each specialized head is strategically mapped to distinct feature levels of the InternVideo2~\cite{wang2024internvideo2} backbone. 
The physical head ingests low-level, early-layer features to detect fine-grained violations, while the task head ingests deep, late-layer features to evaluate high-level semantic completion.

Furthermore, to overcome the \emph{Algorithm Barrier}, we introduce HERO-FPO (HERO-guided Flow Policy Optimization), which operationalizes Flow Policy Optimization (FPO)~\cite{mcallister2025flow} for aligning high-resolution flow-matching video models. Standard PPO is intractable for flow models because computing $\log \pi_\theta$ (and the associated likelihood-ratio term) requires an $\mathcal{O}(d^2 \cdot T_{ODE})$ Jacobian-trace integration. HERO-FPO instead uses a tractable surrogate based on the Conditional Flow Matching (CFM) loss to approximate the likelihood term needed for PPO-style updates, reducing the update complexity to $\mathcal{O}(d)$ and enabling feasible RLHF-style optimization for flow-based video generators. Our contributions are summarized as follows:
\begin{itemize}[leftmargin=1.2em,topsep=2pt,itemsep=2pt]
\item We introduce ReWorld, a novel framework that improves physical realism, embodiment consistency, task success, and visual fidelity—bridging the long-standing gap between visually plausible and physically grounded embodied video generation models.
\item We collect a large-scale embodied preference-pair dataset and introduce HERO, a hierarchical reward model specialized for enabling fine-grained physics understanding and high-level semantic reasoning.
\item We propose HERO-FPO, which provides a Flow Policy Optimization method for embodied video generation.
\end{itemize}

For further evaluation, we introduce ReWorldBench, a new embodied benchmark specifically designed to quantify failures within the \emph{Physics Uncanny Valley}.
Extensive experiments demonstrate that ReWorld achieves state-of-the-art results on ReWorldBench, with clear gains in the aggregate score and especially in the physical realism and embodiment dimensions, showing that multi-dimensional reward alignment substantially narrows the \textit{Physics Uncanny Valley} gap.\looseness=-1

\section{Methodology}

In this section, we detail ReWorld, a framework aimed to employ RL to align the embodied video generation models with physical realism, task completion capability, embodiment plausibility and visual quality.
As illustrated in Fig.~\ref{fig:teaser}, we first construct a 4-dimensional (4D) video preference dataset, which serves as the foundation for defining ``trustworthiness'' (Sec. \ref{ssec:The 4D Embodied Preference Dataset}).
We then introduce a multi-dimensional reward model trained on the aforementioned dataset (Sec. \ref{ssec:HERO}).
Finally, we present a novel reinforcement learning algorithm to optimize flow-based video generation models (Sec. \ref{ssec:HERO-FPO}).

\subsection{The 4D Embodied Preference Dataset}
\label{ssec:The 4D Embodied Preference Dataset}
To solve the ``\textit{Reward Barrier}'' problem, we first create a training signal that transcends simple, monolithic labels. 
Our primary objective is to formalize the complex and often conflicting principles of embodied interaction into orthogonal evaluation criteria.
To this end, we define the 4-Dimensional embodied preference space $\mathcal{P} = \mathbb{R}^4$:
\begin{enumerate}[leftmargin=1.2em,topsep=2pt,itemsep=2pt]
\item \textbf{Physical Realism ($R_{phys}$):} Adherence to physical laws (\textit{e.g.}, object permanence, collision, gravity).
\item \textbf{Embodiment Plausibility ($R_{embod}$):} Kinematic realism and smoothness of the agent's motion.
\item \textbf{Task Completion ($R_{task}$): }Logical and semantic alignment with the given instruction.
\item \textbf{Visual Quality ($R_{vis}$): }Standard visual fidelity, including photorealism and temporal coherence.
\end{enumerate}

To overcome the data scale bottleneck in RLHF, we employ a scalable VLM-driven annotation system. Specifically, we leverage the robust time-series understanding and structural output stability of GPT-4o~\cite{openai2024gpt4technicalreport} as a high-fidelity proxy for human evaluation. 
The VLM is prompted using a structured, information-theoretic template (Appendix~\ref{sec:supp_dataset}) to produce a 4D score vector $\mathbf{s} = [s_{phys}, s_{task}, s_{embod}, s_{vis}] \in [1, 6]^4$ for each video in the RH20T~\cite{Fang2023arXiv} dataset. This approach enables the scalable generation of a $235K+$ preference dataset, a scale sufficient to overcome the data scarcity bottleneck that has hindered previous reward modeling efforts.

Central to our approach is the dimension isolation strategy, which is grounded in the mutual information minimization principle. 
Rather than random pairing, our specialized sampler algorithmically searches for video pairs $(v_A, v_B)$ that serve as highly specific training signals. 
The pair selection is formulated as a constrained combinatorial optimization: maximize the score difference in one target dimension $k$ (\textit{i.e.}, $|s_A^k-s_B^k| > \tau$) while simultaneously minimizing the score difference in all other non-target dimensions $l \neq k$ (\textit{i.e.}, $|s_A^l-s_B^l| < \epsilon$).
The resulting dataset $\mathcal{D} = \{(v_A, v_B, k)\}$ provides a dimensional tag $k$, which serves as the crucial decoupling signal, which enables the functionally specialized and interference-free learning of our reward model in the next stage.

\subsection{HERO: HiErarchical Reward mOdel}
\label{ssec:HERO}
HERO is the multi-dimensional reward model to learn the preferences from our 4D dataset, which is designed to solve the ``\textit{Reward Barrier}'' by our decoupled, multi-dimensional principle. 
It is built upon InternVideo2~\cite{wang2024internvideo2} and features a novel hierarchical reward awareness that strategically maps specialized reward heads to the backbone's feature space.
While a strong VLM is used to generate supervision for dataset construction, we do not rely on invoking the VLM online during RL.
Instead, HERO distills the VLM-generated multi-criteria judgments into a single deployable video reward model, which yields two practical advantages: (i) it enables high-throughput, low-variance reward computation on large batches of generated rollouts, and (ii) it makes the reward stationary across optimization iterations, avoiding prompt- and format-sensitivity that can introduce non-stationarity.

As detailed in Fig.~\ref{fig:hero_architecture}(b), HERO employs four decoupled reward heads: $R_{phys}$, $R_{embod}$, $R_{task}$, and $R_{vis}$, each specializing in one of the 4D dimensions defined in Sec.~\ref{ssec:The 4D Embodied Preference Dataset}. 
This is crucial because it aligns the perceptual complexity required for each evaluation dimension with the corresponding semantic depth of the backbone's features.

The training objective for HERO should solve two distinct challenges simultaneously. 
First, we must enforce that each specialized head learns only its assigned dimension, preventing gradient interference from the others. 
Second, we must ensure all four heads output scores on a comparable numerical scale.
To solve both challenges, we introduce two specialized loss components. 
The dimensional specificity loss ($\mathcal{L}_{D}$), detailed in Sec.~\ref{ssec:The Dimensional Specificity Loss}, addresses the first challenge by utilizing our dataset's dimensional tags to train each head in isolation. 
The overall preference regularizer loss ($\mathcal{L}_{O}$), detailed in Sec.~\ref{ssec:The Overall Preference Regularizer}, then addresses the second challenge by applying a loss to the final combined score, forcing all specialized heads onto a calibrated numerical scale.

\subsubsection{The Dimensional Specificity Loss}
\label{ssec:The Dimensional Specificity Loss}
The dimensional specificity loss $\mathcal{L}_{D}$ is the core objective that enforces functional specialization. 
It leverages the dimension-tagged pairs $(v_A, v_B, k)$ from our dataset to train each head only on its relevant preference signal.
It is computed as a weighted and masked sum of the standard Bradley-Terry loss ($\mathcal{L}_{BT}$) across the four dimensions:

{\small
\begin{align}
&\mathcal{L}_{D}=\sum_{k=1}^4 \mathbb{E}_{(v_A, v_B, k) \sim \mathcal{D}} \left[ \mathbf{W}_{k} \cdot \mathbf{M}_{k} \cdot \mathcal{L}_{BT}(R_k(v_A), R_k(v_B)) \right].
\label{eq:MW}
\\
&\mathcal{L}_{BT}(R_k(v_A), R_k(v_B))=-\log(\sigma(R_k(v_A)-R_k(v_B))).
\end{align}
}

Here, the dimensional mask $\mathbf{M}_k\in\{0,1\}$ acts as a gradient gate that preserves the purity of the supervision: it allows updates only when the preference in the target dimension is sufficiently unambiguous (i.e., the teacher score difference exceeds a margin $\tau$), which prevents the head from learning from entangled or low-confidence comparisons.
The adaptive weight $\mathbf{W}_k$ further improves sample efficiency by prioritizing comparisons with larger teacher preference margins (e.g., scaling with $|\Delta s_k|$) and by upweighting explicitly dimension-isolated pairs produced by our sampler in Sec.~\ref{ssec:The 4D Embodied Preference Dataset}. 
Together, $\mathbf{M}_k$ and $\mathbf{W}_k$ concentrate gradient updates on informative and dimension-specific signals, enabling each head to converge to a specialized decision boundary.

\subsubsection{The Overall Preference Regularizer}
\label{ssec:The Overall Preference Regularizer}
The $\mathcal{L}_{D}$ objective successfully trains specialized heads, but it does not guarantee that their output scores are calibrated with one another. For example, $R_{phys}$ might learn to output scores in the range [0.1, 0.2] while $R_{task}$ outputs in [1.0, 5.0]. If combined, $R_{task}$ would completely dominate the final reward. 
The overall preference regularization $\mathcal{L}_{O}$ solves this score calibration problem. Instead of acting on individual heads, it is a Bradley-Terry loss computed on the final, combined scalar reward $R_{total} = \sum_{k} w_k R_k$:
\begin{align}
\mathcal{L}_{O} = \mathcal{L}_{BT}(R_{total}(v_A), R_{total}(v_B)).
\end{align}
This loss serves as a general quality constraint, forcing all four specialized heads to learn to produce scores within a comparable and meaningful numerical range. 

Finally, the total loss $\mathcal{L}_{HERO}$ is formulated as a composite objective, combining both solutions:
\begin{equation}
\mathcal{L}_{HERO} = \beta \cdot \mathcal{L}_{D} + (1 - \beta) \cdot \mathcal{L}_{O},
\end{equation}
where $\beta$ heavily emphasizes specialization, while $\mathcal{L}_{O}$ provides robust calibration.
The resulting HERO model is frozen upon convergence, and its combined, calibrated scalar output $R = \sum_{k} w_k R_k$ is used as the high-fidelity reward function in the HERO-FPO pipeline (\ref{ssec:HERO-FPO}).

\subsection{HERO-FPO: HERO-guided Flow Policy Optimization}
\label{ssec:HERO-FPO}
With HERO fixed as a calibrated reward function $R(v,c)$, we optimize a conditional flow-matching video generator $\pi_\theta(v\mid c)$ (Cosmos-2B/14B~\cite{alhaija2025cosmos}) via reinforcement learning, where $c$ contains the initial observation and the instruction.
Because the general-domain prior exhibits a strong domain gap for contact-rich embodied interactions, we first supervise fine-tune Cosmos on Bridge V2~\cite{walke2023bridgedata} with the standard CFM objective, obtaining a domain-adapted initialization for the RL stage.

We then perform PPO-style optimization: rollouts are sampled from $\pi_{\theta_{old}}$, scored by the frozen reward model, and converted to advantages using a learned critic, after which the actor and critic are updated with a clipped objective.
The central difficulty is that PPO requires the importance ratio based on $\log \pi_\theta(v\mid c)$, which is intractable for high-dimensional flow-based video models.
HERO-FPO resolves this by replacing the likelihood ratio with a tractable surrogate derived from CFM losses, enabling stable PPO-style updates for flow-based video generation.

{
\begin{algorithm}[t]
\small
\caption{HERO-FPO}
\label{alg:HERO-FPO}
\begin{algorithmic}[1]
\setlength{\itemsep}{0pt}
\setlength{\parsep}{0pt}

\State \textbf{Initialize:} Actor policy $\pi_\theta$ (Cosmos), Critic $V_\psi$, frozen Reward Model $R_{HERO}$, Old policy $\pi_{\theta_{old}} \gets \pi_\theta$.

\Statex \textbf{Stage 1: Experience Collection}

\For{iteration $i = 1, \dots, N$}

    \State Sample batch of conditions $\mathcal{C} = \{c_j\}_{j=1}^B$ from RH20T.

    \For{each $c_j \in \mathcal{C}$}
        \State Generate video $v_j \sim \pi_{\theta_{old}}(v \mid c_j)$ (using Cosmos).
        \State Compute 4D rewards $\mathbf{R}_{4D} \gets R_{HERO}(v_j, c_j)$ (using HERO).
        \State Compute total scalar reward $R_j \gets \sum_{k} w_k R_k$.
        \State Compute value $V_j \gets V_\psi(v_j)$.
        \State Compute advantage $\hat{A}_j \gets R_j - V_j$.
        \State Store trajectory $\mathcal{T}_j = (v_j, c_j, R_j, \hat{A}_j)$ into buffer $\mathcal{B}$.
    \EndFor

    \Statex \textbf{Stage 2: FPO Optimization}

    \For{epoch $e = 1, \dots, K$}
        \For{$(v, c, R, \hat{A}) \in \mathcal{B}$}

            \State {\fontsize{7pt}{8pt}\selectfont \textit{\# Compute ratio using CFM-Likelihood Proxy (\ref{eq:cfm_surrogate})}}

            \State $r(\theta)
            \gets \exp\!\left(
                L_{CFM}(v;\theta_{old},c)
                - L_{CFM}(v;\theta,c)
            \right)$

            \State {\fontsize{7pt}{8pt}\selectfont 
    $L_{\text{POLICY}}(\theta)
        \gets -\min\!\Big(
            r(\theta)\hat{A},\;
            \text{clip}(r(\theta),1{-}\epsilon,1{+}\epsilon)\hat{A}
        \Big)$
}

            \State $L_{\text{VALUE}}(\psi)
                \gets (R - V_\psi(v))^2$

            \State $L_{\text{TOTAL}}
                \gets L_{\text{POLICY}} + c_v L_{\text{VALUE}}$

            \State Update $\theta, \psi$ via $\nabla L_{\text{TOTAL}}$

        \EndFor
    \EndFor

    \State $\pi_{\theta_{old}} \gets \pi_\theta$

\EndFor

\end{algorithmic}
\end{algorithm}
}

\normalsize\selectfont
\subsubsection{The CFM-Likelihood Proxy}
The foundational challenge of applying PPO to a flow-based model $\pi_\theta$ is the ``\textit{Algorithm Barrier}'': the calculation of the log-likelihood, $\log \pi_\theta(v|c)$, is computationally intractable for high-dimensional video, as it requires integrating the trace of the model's Jacobian $f_\theta$:
{
\begin{align}
\log \pi_\theta(v|c) ^= \log p_0(z_0) - \int_0^1 \nabla \cdot f_\theta(z_t, t) dt.
\\
(\text{Intractable: } \mathcal{O}(d^2 \cdot T_{ODE})) . 
\end{align}
}
To establish our proxy, we first formally define the $L_{CFM}$ objective. The CFM loss, used to train the base model $\pi_\theta$, is a tractable MSE objective. It is computed by taking the clean video $v$, adding a random amount of noise $\epsilon \sim \mathcal{N}(0, I)$ at a random noise level $\sigma$, creating a noised video $v_t = v + \sigma\epsilon$. The model $\pi_\theta$ is then tasked with predicting the original clean video $v$ given $v_t$, $\sigma$, and $c$. The loss is the weighted squared error between the prediction and the ground truth:

{\small
\begin{align}
L_{CFM}(v; \theta, c) = \mathbb{E}_{\sigma, \epsilon} \left[ w(\sigma) \cdot \| \pi_\theta(v_t, \sigma, c) - v \|^2 \right].
\label{eq:cfm_surrogate}
\end{align} 
}
Our key insight, which we term the CFM-Likelihood Proxy, is that this loss value itself—a measure of how well the model can denoise $v$—is a powerful proxy for likelihood. Intuitively, a low $L_{CFM}$ (the model easily denoises $v$) implies that $v$ is high log-likelihood. Conversely, a high $L_{CFM}$ (the model struggles to denoise $v$) implies $v$ is low log-likelihood. This strong negative correlation forms the basis of our proxy:
\begin{align}
\log \pi_\theta(v|c) \approx -L_{CFM}(v; \theta, c) + C(c).
\end{align}
Crucially, the constant term $C(c)$ depends only on the condition $c$ and cancels out during the PPO update. This substitution allows us to compute the PPO importance sampling ratio $r(\theta) = \pi_\theta / \pi_{\theta_{old}}$ entirely in terms of tractable CFM losses. The log-ratio simplifies dramatically:
\begin{align}
\log r(\theta) &= \log \pi_\theta - \log \pi_{old} \\
&\approx \left(-L_{CFM}^{new} + C\right) - \left(-L_{CFM}^{old} + C\right) .
\end{align}
This yields our final, computationally feasible update rule:
\begin{equation}
r(\theta) \approx \exp\left(L_{CFM}^{old}(v; \theta_{old}, c) - L_{CFM}^{new}(v; \theta, c)\right).
\end{equation}
This theoretical breakthrough is the core of HERO-FPO. It reduces the complexity of the PPO update from the impossible $\mathcal{O}(d^2 \cdot T_{\text{ODE}})$ to a highly efficient $\mathcal{O}(d)$, making high-resolution video RLHF computationally feasible.

\subsubsection{The HERO-FPO-PPO Training Framework}
Building on the CFM-Likelihood Proxy in~\ref{eq:cfm_surrogate}, we instantiate a fully tractable actor--critic PPO training loop for aligning flow-based conditional video generation.
This choice differs from recent GRPO-style optimization for visual generators (e.g., DanceGRPO~\cite{xue2025dancegrpo}), which constructs group-relative advantages by normalizing rewards across multiple samples under the same condition, and applies a clipped-ratio update with respect to the previous policy. In practice, this group-relative formulation avoids training an explicit value critic, whereas our actor--critic PPO relies on a learned baseline and, critically, requires a tractable likelihood ratio for flow-based video policies, which is enabled by our CFM-Likelihood Proxy.

In contrast, PPO relies on the likelihood ratio $r(\theta)=\pi_\theta(v\mid c)/\pi_{\theta_{old}}(v\mid c)$ and benefits from a learned value baseline to reduce variance, which is particularly important when rewards emphasize subtle physical and embodied correctness.

The core of the optimization (Stage 2 in~\ref{alg:HERO-FPO}) uses the PPO clipped objective to update the Actor and Critic simultaneously:
{\small
\begin{align}
L_{\text{POLICY}}(\theta)=-\min\left(r(\theta)\hat{A}, \text{clip}(r(\theta), 1-\epsilon, 1+\epsilon)\hat{A}\right).
\end{align}
}
\label{sec:models}

\begin{table*}[t!]
\label{tab:standard_video_metrics}
\centering
\caption{\textbf{Quantitative comparison on ReWorldBench.} The evaluation covers both visual fidelity and performance on embodied tasks. We highlight the best results in \textbf{bold}.}
\resizebox{\columnwidth}{!}{%
\begin{tabular}{@{}l|ccccc|cccc|c@{}}
\toprule
& \multicolumn{5}{c}{\textbf{Video Quality}} & \multicolumn{4}{c}{\textbf{4D Embodied Dimensions (1-10 Scale)}} & \textbf{Overall}\\
\cmidrule(lr){2-6} \cmidrule(lr){7-10} \cmidrule(lr){11-11}
\textbf{Model} & \textbf{FVD} ($\downarrow$) & \textbf{SSIM} ($\uparrow$) & \textbf{DINO Similarity} ($\uparrow$) & \textbf{PSNR} ($\uparrow$) & \textbf{DreamSim} ($\uparrow$) & \textbf{$S_{phys}$} ($\uparrow$) & \textbf{$S_{embod}$} ($\uparrow$) & \textbf{$S_{task}$} ($\uparrow$) & \textbf{$S_{vis}$} ($\uparrow$) & \textbf{$S_{ReWorld}$} ($\uparrow$)\\
\midrule
CogVideoX~\cite{yang2025cogvideoxtexttovideodiffusionmodels} & 720 & \textbf{0.70} & 0.56 & 14.6 & 0.74 & 3.1 & 2.8 & 3.4 & 7.1 & 35.3\\
Wan2.1~\cite{wan2025wanopenadvancedlargescale} & 510 & 0.61 & 0.59 & 14.1 & 0.74 & 3.4 & 3.1 & 3.8 & 7.3 & 38.6\\
Cosmos-Base~\cite{alhaija2025cosmos} & 281 & 0.63 & 0.62 & 15 & 0.74 & 4.2 & 3.9 & 4.4 & 7.0 & 44.7\\
Cosmos-SFT (Ours) & 240 & 0.68 & 0.65 & \textbf{16.1} & 0.78 & 5.1 & 4.2 & 6.1 & 7.2 & 54.4\\
\rowcolor{linecolor2}{\textbf{ReWorld (Ours)}} & \textbf{190} & 0.66 & \textbf{0.71} & 15.2 & \textbf{0.82} & \textbf{5.9} & \textbf{5.6} & \textbf{6.5} & \textbf{7.3} & \textbf{61.9}\\
\bottomrule
\end{tabular}
}

\end{table*}

\begin{table*}[h]
\caption{\textbf{HERO reward model performance.}}
\label{tab:hero_results}
\centering
\resizebox{\columnwidth}{!}{%
\setlength{\tabcolsep}{20pt}
\begin{tabular}{@{}lc|cccc@{}}
\toprule
\multicolumn{2}{c|}{\textbf{Overall Performance}} & \multicolumn{4}{c}{\textbf{Per-Dimension Accuracy (Specialization)}} \\
\midrule
Metric & Value & Head & Accuracy & Feature Source & Rank \\
\midrule
Accuracy & \textbf{85.3\%} & $R_{embod}$ & \textbf{87.2\%} & Mid-layer (L16) & 1 (Best) \\
AUC & \textbf{0.901} & $R_{vis}$ & 84.6\% & Final-pooled (L39) & 2 \\
Spearman ($\rho$) & \textbf{0.787} & $R_{task}$ & 81.3\% & Deep-layer (L32) & 3 \\
Kendall's ($\tau$) & \textbf{0.689} & $R_{phys}$ & 79.1\% & Early-layer (L8) & 4 (Hardest) \\
\bottomrule
\end{tabular}%
}

\end{table*}
\subsection{ReWorldBench: A Multi-Dimensional Benchmark for Embodied Reality}
\label{ssec:ReWorld-Bench: A Benchmark for Embodied Reality}
In this section, we introduce ReWorldBench, a specialized benchmark designed to evaluate embodied video generation models. We define the core evaluation task as conditional video generation from an initial image and a text instruction (Image+Text-to-Video), a setup that directly probes the model's ability to understand a given state and execute a specified action. ReWorldBench not only provides metrics for visual quality but is also specifically designed to evaluate a model's physical realism, task completion, and embodiment plausibility in embodied settings.

\subsubsection{Evaluation Dimensions}
\label{ssec:Evaluation Dimensions}
\begin{itemize}[leftmargin=1.2em,topsep=2pt,itemsep=2pt]
\item \textbf{Predictive Physical Reasoning (Probing $R_{phys}$):} Evaluates the model's adherence to core physical principles. The videos are assessed for violations in object permanence, collision dynamics and gravitational realism, focusing on dynamic, contact-rich events.
\item \textbf{Logical and Task Planning (Probing $R_{task}$):} Evaluates the model's logical and semantic adherence to the given instruction. Success is measured by whether the generated video performs the correct actions in the specified logical sequence, especially for complex or multi-step tasks.
\item \textbf{Kinematic Execution (Probing $R_{embod}$):} Distinct from world physics, this evaluates the agent's own motion realism. We assess the generated trajectories for kinematic correctness, smoothness, and continuity.
\item \textbf{Generative Fidelity (Probing $R_{vis}$):} Evaluates the foundational generative quality of the model. We assess standard criteria: photorealism, absence of visual artifacts, and temporal consistency.
\end{itemize}
\subsubsection{Task Design and Data Curation}
The benchmark's prompts are built upon the diverse scenarios within the RH20T~\cite{Fang2023arXiv} task set. To specifically probe the four dimensions, we leverage GPT-4o to systematically curate and expand the original instructions.
We will provide the detailed illustrations in the supplementary material.

\subsubsection{Evaluation Protocol}
We employ a rigorous VLM-as-judge protocol built upon GPT-4o~\cite{openai2024gpt4technicalreport}, chosen for its advanced spatio-temporal and fine-grained reasoning capabilities. The core of our protocol is a set of dimension-specific, rationale-first evaluation templates. For each (video, prompt) pair, the template elicits a detailed textual rationale prior to requesting a numerical score (1--10) for each of the four dimensions. This rationale-first approach ensures the evaluation is interpretable, consistent, and grounded in specific evidence, mitigating the common VLM biases of ungrounded, holistic scoring.

\section{Experiments}
Our evaluation of the ReWorld framework is structured into three primary analyses. We first detail our comprehensive experimental setup in Sec.~\ref{ssec:Experimental Setup}. \ref{ssec:Main Experimental Results and Analysis} then presents our main quantitative and qualitative results; this section begins by demonstrating the superiority of our full HERO-FPO pipeline against SOTA baselines (Sec. \ref{ssec:World Model Alignment Performance}) and subsequently validates the efficacy of the HERO reward model that enabled this alignment (Sec. \ref{ssec:HERO Reward Model Performance}). Finally, Sec. \ref{ssec:Ablation Study} provides comprehensive ablation studies to isolate and justify our critical design components.

\subsection{Experimental Setup}
\vspace{-1mm}
\label{ssec:Experimental Setup}
Our policy alignment is conducted on robotics tasks from the RH20T dataset~\cite{Fang2023arXiv}, using provided initial condition frames. The reward function is our pre-trained, frozen HERO model, which is built on an InternVideo2-1B backbone~\cite{wang2024internvideo2}. HERO itself is trained on our novel 4D Embodied Preference Dataset ($\sim$235K pairs), which we generated from RH20T using GPT-4o~\cite{openai2024gpt4technicalreport}. The policy model is the Cosmos-2B~\cite{alhaija2025cosmos}, which we pre-finetune on the Bridge V2 dataset~\cite{walke2023bridgedata} to create our strong Cosmos-SFT baseline. The Critic ($V_\psi$) is a 4-layer 3D-CNN followed by a 2-layer MLP VideoValueNetwork trained from scratch. We compare our full ReWorld framework against Cosmos-SFT and other SOTA baselines, including CogVideoX~\cite{yang2025cogvideoxtexttovideodiffusionmodels}, Wan2.1~\cite{wan2025wanopenadvancedlargescale}, and the original Cosmos-Base~\cite{alhaija2025cosmos}.

All models are trained on 8 NVIDIA A100 GPUs using the AdamW optimizer. Detailed hyperparameters for both HERO training and HERO-FPO alignment, are provided in the supplementary material. Evaluation is two-fold: (i) We assess HERO's Accuracy, AUC, Spearman, and Kendall's on the test set. (ii) We evaluate the generative models using standard metrics (FVD~\cite{unterthiner2019accurategenerativemodelsvideo}, SSIM~\cite{wang2004ssim}, DINO Similarity~\cite{caron2021emergingpropertiesselfsupervisedvision}, PSNR, DreamSim~\cite{fu2023dreamsim}) and our proposed ReWorld-Bench (\ref{ssec:ReWorld-Bench: A Benchmark for Embodied Reality}).

\subsection{Main Experimental Results and Analysis}
\label{ssec:Main Experimental Results and Analysis}

\subsubsection{Video Generation Alignment Performance}
\label{ssec:World Model Alignment Performance}
Table~\ref{tab:standard_video_metrics} presents our comprehensive comparison, benchmarking our full framework against all baselines across both standard visual quality metrics and our rigorous 4D ReWorldBench dimensions. As shown in Table~\ref{tab:standard_video_metrics}, ReWorld remains competitive in standard visual metrics, confirming that our HERO-FPO alignment refines embodied behavior without sacrificing foundational visual fidelity.

However, these standard metrics are physics-agnostic and fail to capture the \textit{Physics Uncanny Valley}. 
We thus evaluate all models on our rigorous ReWorldBench (Sec. \ref{ssec:ReWorld-Bench: A Benchmark for Embodied Reality}), which probes the four dimensions of embodied intelligence. The results in Table~\ref{tab:standard_video_metrics} are definitive. While baselines (including our strong Cosmos-SFT) struggle with physical and logical coherence, our full ReWorld framework achieves dramatic improvements across all four embodied dimensions, especially in $S_{phys}$ and $S_{task}$. This provides strong quantitative evidence that our HERO-FPO pipeline successfully closes the \textit{Physics Uncanny Valley}.

\begin{figure}[!t]
  \centering
   \vspace{-7mm}
   \includegraphics[width=1.0\linewidth]{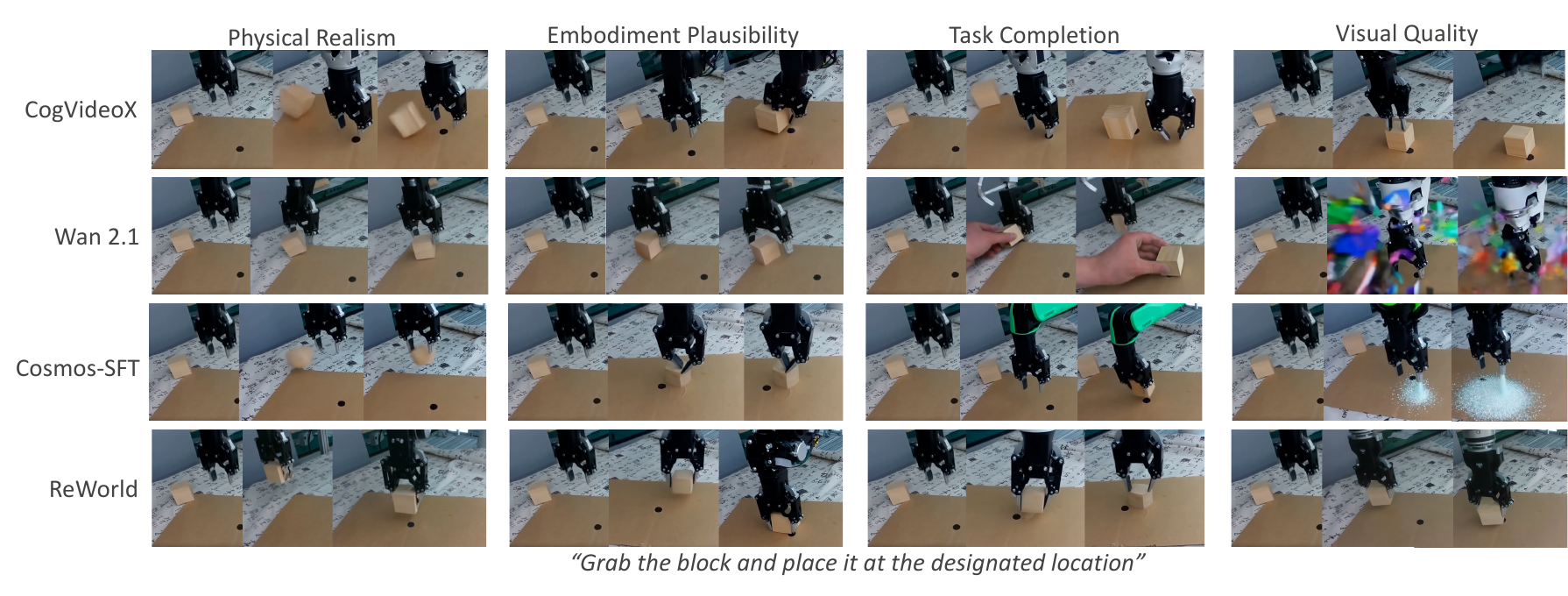}
   \vspace{-4mm}
   \caption{\textbf{Qualitative comparisons on ReWorldBench.} Our proposed ReWorld model achieves the best generative results for all the multi-dimensional metrics compared with the baseline video generation models.}
   \label{fig:visualization}
\end{figure}

\textbf{Qualitative Results.} We provide qualitative comparisons in Fig~\ref{fig:visualization}. The visualizations clearly demonstrate the failures of baseline models, which directly correspond to their low $S_{ReWorld}$ scores. Baselines exhibit severe physical implausibility (CogVideoX), catastrophic visual artifacts and task failure (Wan 2.1), or unnatural kinematics (Cosmos-SFT). In stark contrast, our full ReWorld framework successfully generates a video that is physically, kinematically, and logically coherent, visually confirming its superior $S_{ReWorld}$ score.

\vspace{-1mm}

\vspace{-1mm}
\subsubsection{HERO Reward Model Performance}
\label{ssec:HERO Reward Model Performance}
To rigorously evaluate our HERO and validate its alignment with true human judgment, we constructed an expert-annotated test set. We sampled 300 videos from RH20T~\cite{Fang2023arXiv} dataset and had them meticulously annotated by 5 expert human volunteers, who used the exact 4D evaluation protocol (Sec. \ref{ssec:Evaluation Dimensions}) to create a high-quality, human-validated preference dataset.

As shown in Table~\ref{tab:hero_results}, HERO achieves outstanding performance on this challenging expert-annotated test set. The results confirm HERO learns a strong, generalizable preference signal that successfully transfers from our large-scale VLM training data. Furthermore, the model demonstrates clear functional specialization, with all four heads achieving high accuracy. This validates our hierarchical design (Sec. \ref{ssec:HERO}) and proves its capability as a reliable, multi-dimensional reward function for the FPO alignment stage.

\vspace{-1mm}
\subsection{Ablation Study}
\label{ssec:Ablation Study}

\vspace{-2mm}
\begin{table}[H]
\centering
\begin{minipage}[t]{0.49\textwidth}
\centering
\caption{\textbf{HERO reward model ablation.}}
\label{tab:hero_ablation}
\scriptsize
\renewcommand{\arraystretch}{1}
\resizebox{\linewidth}{!}{%
\setlength{\tabcolsep}{5pt}
\begin{tabular}{@{}lcc@{}}
\toprule
\textbf{Model Variant} & \textbf{Acc.} & \textbf{Drop} \\
\midrule
\textbf{HERO (Full Model)} & \textbf{85.3\%} & \textbf{--} \\
\midrule
\multicolumn{3}{@{}l}{\textit{A. Removing Reward Heads:}} \\
\quad w/o $R_{task}$ (Alignment) & 70.1\% & -15.2\% \\
\quad w/o $R_{embod}$ (Execution) & 73.5\% & -11.8\% \\
\quad w/o $R_{phys}$ (Plausibility) & 75.8\% & -9.5\% \\
\quad w/o $R_{vis}$ (Fidelity) & 79.2\% & -6.1\% \\
\midrule
\multicolumn{3}{@{}l}{\textit{B. Loss Function Components:}} \\
\quad w/o $\mathcal{L}_{D}$ (only $\mathcal{L}_{O}$) & 65.4\% & -19.9\% \\
\quad w/o $\mathcal{L}_{O}$ (only $\mathcal{L}_{D}$) & 75.1\% & -10.2\% \\
\quad $\mathcal{L}_{HERO}$ (Equal Weights) & 81.7\% & -3.6\% \\
\midrule
\multicolumn{3}{@{}l}{\textit{C. Feature Mapping:}} \\
\quad w/o Hierarchical Map & 72.8\% & -12.5\% \\
\bottomrule
\end{tabular}
}
\end{minipage}
\hfill
\begin{minipage}[t]{0.49\textwidth}
\centering
\caption{\textbf{HERO-FPO alignment ablation.}}
\label{tab:fpo_ablation}
\scriptsize
\setlength{\tabcolsep}{2.2pt}
\renewcommand{\arraystretch}{1}
\resizebox{\linewidth}{!}{%
\begin{tabular}{@{}lcc@{}}
\toprule
\textbf{FPO Variant} & $\boldsymbol{S_{ReWorld}}$ & \textbf{Drop} \\
\midrule
\textbf{ReWorld (Full Model)} & \textbf{61.9} & \textbf{--} \\
\midrule
\multicolumn{3}{@{}l}{\textit{A. Core Algorithm:}} \\
\quad w/o CFM-Likelihood Surrogate & 55.1 & -6.8 \\
\quad w/ PPO, wrong sign & 37.8 & -24.1 \\
\midrule
\multicolumn{3}{@{}l}{\textit{B. Reward Components:}} \\
\quad only $R_{phys}+R_{embod}$ & 49.2 & -12.7 \\
\quad only $R_{task}+R_{vis}$ & 51.3 & -10.6 \\
\quad equal reward weights & 58.2 & -3.7 \\
\midrule
\multicolumn{3}{@{}l}{\textit{C. CFM Sampler:}} \\
\quad CFM ($N=1$) & 56.9 & -5.0 \\
\quad CFM ($N=10$) & 60.8 & -1.1 \\
\bottomrule
\end{tabular}
}
\end{minipage}
\vspace{-2mm}
\end{table}

\noindent
\textbf{HERO Reward Model.} Table~\ref{tab:hero_ablation} presents the ablation results for the core components of the HERO. The results indicate that all three design pillars are essential for performance. The most critical component is our Dimensional Specificity Loss ($\mathcal{L}_{D}$), as removing it causes a catastrophic 19.9\% drop in accuracy. 
Our \textit{Hierarchical Reward Awareness} hypothesis (Sec.~\ref{ssec:HERO}) is also strongly validated, as reverting to a naive final-layer feature map degrades performance by 12.5\%. Furthermore, all four heads prove to be necessary, with the $R_{task}$ head being the most impactful, whose removal results in a 15.2\% accuracy loss. These results validate that HERO's high performance stems from the synergistic design of its hierarchical architecture and its specialized, composite loss function.

\noindent
\textbf{HERO-FPO.} Table~\ref{tab:fpo_ablation} presents the ablation results for our HERO-FPO alignment framework. The results confirm that the CFM-Likelihood Proxy (\ref{ssec:HERO-FPO}) is the most critical component. Replacing this proxy with a naive reward-weighted $L_2$ loss, or incorrectly using the $L_{CFM}$ value as a direct $\log \pi$ proxy, causes a catastrophic performance collapse of 24.1 points. This validates that our principled proxy (\ref{eq:cfm_surrogate}) is essential for stable policy optimization. Furthermore, the multi-dimensional reward signal is proven vital; ablating the physics and embodiment heads results in a (-12.7 points) loss, confirming these dimensions are key to closing the \textit{Physics Uncanny Valley}. Finally, our choice of $N=5$ samples for the CFM sampler is validated as a robust trade-off, as using only $N=1$ sample degrades performance by 5 points due to high variance, while increasing to $N=10$ offers no significant benefit.

\vspace{-2mm}

\label{sec:experiments}

\section{Conclusion and Future Work}
\vspace{-2mm}
In this paper, we propose ReWorld, a new framework for aligning embodied video generation models by systemically resolving the core reward and algorithm barriers in video RLHF. 
It introduces HERO, a multi-dimensional reward model with hierarchical reward awareness to solve the reward challenge, and enables HERO-FPO, a tractable PPO algorithm grounded in our CFM-likelihood proxy theory to solve the optimization challenge. 
This integration delivers state-of-the-art performance compared with previous methods.\looseness=-1

\noindent

\label{sec:conclusion}

\bibliographystyle{plainnat}
\bibliography{main.bib}

\newpage
\appendix


\section*{\huge \centering Appendix}

\startcontents
{
    \hypersetup{linkcolor=black}
    \printcontents{}{1}{}
}

\newpage

\section{Related Works}
\subsection{Embodied Video Generation and World Models}

Embodied World Models (EWMs)~\cite{ha2018world, B_planet2019, C_dreamer2020, D_muzero2020, hafner2023mastering} learn latent dynamics for planning and policy learning. 
In parallel, large-scale video generation has rapidly advanced~\cite{sora2024, villegas2022phenaki, singer2022makeavideo, ho2022video}, enabling high-fidelity and increasingly general video/world simulators~\cite{liu2024worldmodel, hu2023gaia1, li2025omninwm}, including flow-based generators such as Cosmos~\cite{alhaija2025cosmos}. Controllable video generation further improves alignment to conditions like text, motion, or depth~\cite{zhang2023controlnet, zhang2023controlvideo, esser2023structure, wang2023videocomposer, wu2023tuneavideo, chen2023control, yang2023rerender}.

Despite their visual quality, many video generators and EWMs are trained primarily with pixel-level reconstruction losses (\textit{e.g.}, $L_2$, $L_1$, or LPIPS~\cite{zhang2018unreasonable})~\cite{ha2018world, alhaija2025cosmos}. Such objectives encourage perceptual similarity but are physics-agnostic, providing weak supervision for physical plausibility, contact correctness, and long-horizon task logic. As a result, even strong conditional models~\cite{wang2023videocomposer, chen2023control, guo2023animatediff, wang2024motionctrl, wang2024magicvideov2} may produce visually plausible yet physically inconsistent rollouts. ReWorld addresses this by introducing a post-hoc RL alignment pipeline that optimizes embodied rollout policies with multi-dimensional rewards beyond pixel reconstruction.

\subsection{Policy Optimization for Generative Models}

RL has been used to optimize generative models such as GANs and VAEs~\cite{goodfellow2014generative, B_seqgan2017, kingma2013auto}. Recent work successfully applies RLHF-style alignment to diffusion models~\cite{black2023training, lee2023aligning, ren2024diffusion, C_imagereward2023}, leveraging the tractable noise-prediction objective as a proxy for likelihood~\cite{song2020denoising}. However, flow-based models~\cite{kingma2018glow, song2021score, lipman2022flow, D_consistency2023} are typically trained without explicit log-likelihood, since exact likelihood involves expensive Jacobian terms~\cite{grathwohl2018ffjord, chen2018neural}. Modern generators such as Cosmos~\cite{alhaija2025cosmos} adopt Flow Matching (CFM)~\cite{mcallister2025flow, lipman2022flow, albergo2023stochastic}, which yields a stable MSE objective but does not directly provide $\log \pi_\theta$.
This makes PPO-style optimization~\cite{schulman2017proximal} challenging for flow-matching generators, since it requires likelihood ratios $r(\theta)=\pi_\theta/\pi_{\text{old}}$ and thus $\log \pi_\theta$. Concurrent work explores scalable RL fine-tuning for visual generation, \textit{e.g.}, DanceGRPO~\cite{xue2025dancegrpo}. ReWorld introduces HERO-FPO with a CFM-Likelihood Proxy that connects changes in the tractable CFM loss to a likelihood-ratio-like term for PPO updates, enabling practical PPO-style alignment for flow-matching video models under multi-dimensional embodied rewards.
\label{sec:related_works} 

\subsection{Reward Modeling for Vision and Robotics}

RLHF~\cite{christiano2017deep} has been successful for aligning LLMs~\cite{ouyang2022training, ziegler2019fine} and has inspired reward modeling in vision~\cite{brooks2023instructpix2pix, lee2023aligning}, including rewards for aesthetics~\cite{ye2024dreamreward, schuhmann2022laion} and text-image alignment~\cite{kirstain2023pick, wallace2024diffusion}. These reward models often output a single scalar preference, while robotics commonly uses sparse success signals or hand-crafted heuristics~\cite{chi2025wow, mees2022calvin, hadfield2017inverse, aytar2018playing, tang2025deep}. Neither is well-suited for refining high-dimensional video rollouts: monolithic scores conflate aesthetics with physical correctness~\cite{ye2024dreamreward, radford2021learning}, and sparse rewards provide limited shaping signals~\cite{chi2025wow}.

A related trend uses VLMs as judges to provide scalable, rubric-based supervision, but directly using an off-the-shelf VLM as an online RL reward can be costly and brittle to prompting or generative artifacts. ReWorld follows a distillation view: we first obtain multi-dimensional supervision offline to build preference data, then train a dedicated spatiotemporal reward model (HERO) that is efficient and consistent for RL alignment. HERO further decouples reward dimensions across feature levels, enabling simultaneous evaluation of low-level physics and high-level semantics.

\section{4D Embodied Preference Dataset Details} 
\label{sec:supp_dataset}

\begin{figure*}[t]
  \centering
  \includegraphics[width=1.0\linewidth]{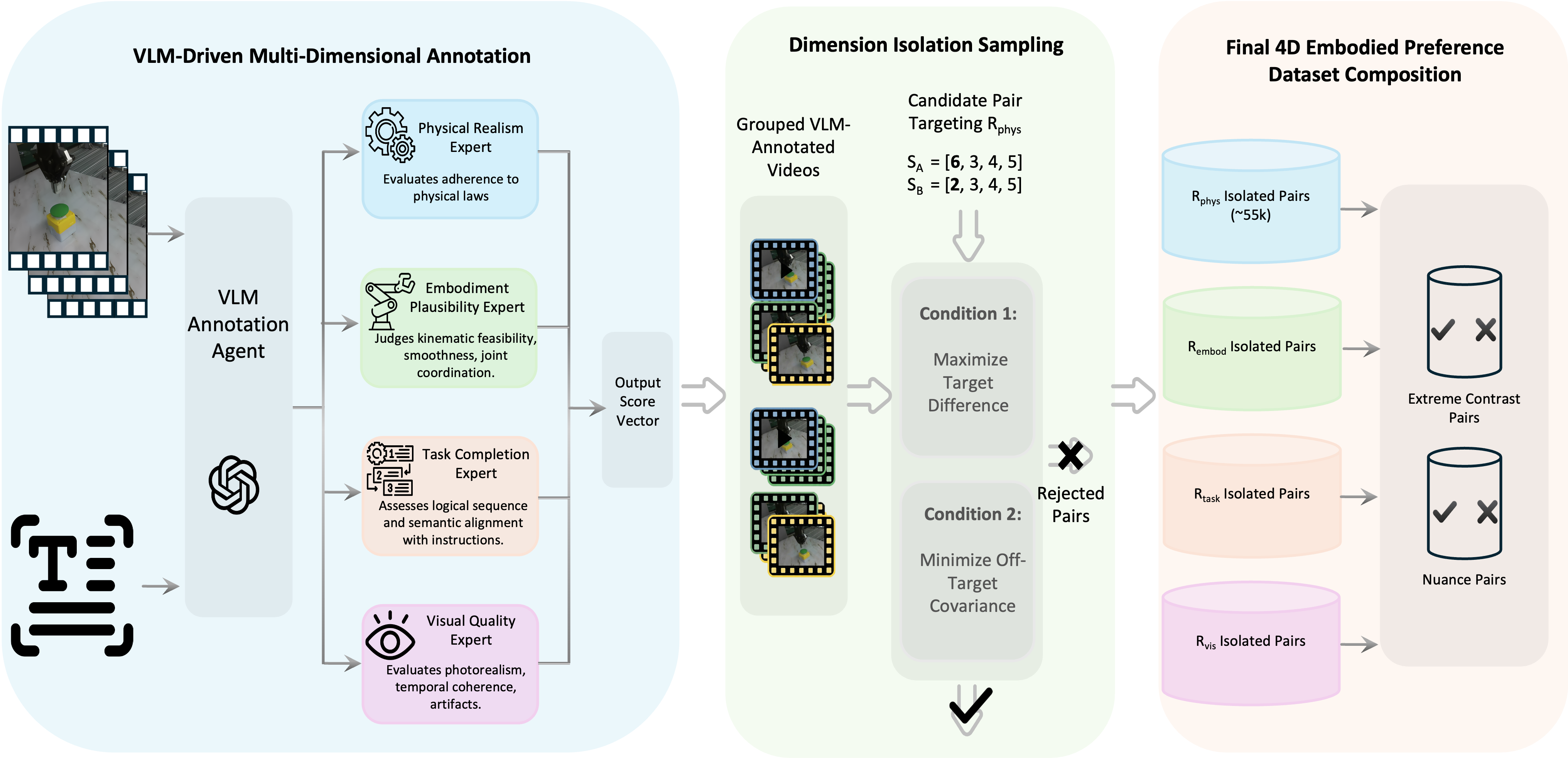} 
  \caption{Pipeline for constructing the 4D Embodied Preference Dataset. 
  The process consists of three stages: 
  (1) \textbf{VLM-Driven Annotation (Left):} We employ GPT-4o with role-playing prompts to act as four specialized experts, generating a dense 4D score vector for each video. 
  (2) \textbf{Dimension Isolation Sampling (Middle):} A constrained optimization algorithm selects video pairs that maximize difference in a \textit{target dimension} (Condition 1) while minimizing variance in others (Condition 2), ensuring signal purity. 
  (3) \textbf{Final Composition (Right):} The resulting dataset contains $\sim$235K pairs, structured into dimension-isolated subsets for specialized learning and contrastive subsets for global calibration.}
  \label{fig:data_pipeline}
\end{figure*}

This section provides a detailed exposition of the 4-Dimensional (4D) Embodied Preference Dataset. The comprehensive construction pipeline is visually summarized in \ref{fig:data_pipeline}. The construction of this dataset is foundational to overcoming the ``\textit{Reward Barrier}'' defined in the main paper. We elucidate the VLM-as-annotator protocol, the dimension-specific sampling algorithms, and the final dataset statistics as depicted in the figure's three stages. 
\subsection{VLM Annotation Protocol}As stated in the main paper, we leverage GPT-4o~\cite{openai2024gpt4technicalreport} as our expert annotator proxy. This choice was predicated on its advanced spatio-temporal reasoning, logical inference, and robust instruction-following capabilities, which are essential for generating high-fidelity preference labels at scale.

As illustrated in the left panel of \ref{fig:data_pipeline}, our annotation protocol is grounded in information-theoretic templates designed to provide clear, quantitative, and decoupled evaluation criteria. Each of the four dimensions ($R_{phys}, R_{task}, R_{embod}, R_{vis}$) has a dedicated prompt. Each prompt consists of three critical components:

\begin{enumerate}
\item Expert Persona (System Role): As visualized by the specific role icons in \ref{fig:data_pipeline}, the VLM is instructed to act as a domain expert (e.g., ``expert in robotics and Newtonian physics'').
\item Quantitative Criteria (Scoring Rubric): A 6-point Likert scale is provided with explicit, quantitative definitions for each score, along with key evaluation criteria to focus the VLM's attention.
\item Strict Output Schema (JSON Enforcement): The VLM is mandatorily constrained to return only a valid JSON object: {``score'': <int>, ``rationale'': ``<string>''}. This ensures 100\% parse-ability and structural consistency across all $\sim$235K annotations.
\end{enumerate}

The precise protocols for each of the four dimensions are detailed in Box 1 through Box 4.

\begin{figure*}[t]
\centering
\begin{minipage}{0.48\textwidth}
\centering
\fbox{%
\begin{minipage}{0.95\linewidth}
\small
\setlength{\parindent}{0pt}

\textbf{Box 1: $R_{\text{phys}}$ (Physical Realism) Evaluation Protocol}

\vspace{0.2cm}
\hrule
\vspace{0.2cm}

\textbf{System Prompt:} You are an expert in robotics and Newtonian physics.
Your task is to evaluate the physical realism of the interactions in the provided video.

\vspace{1.5ex}

\textbf{User Prompt:}\\[0.5ex]
Physical Realism Evaluation\\
Task: \{task\_description\}

\vspace{1.5ex}

\textbf{Scoring Criteria (1--6 Scale):}
\begin{itemize}[itemsep=0.5ex]
    \item \textbf{6:} Flawless. All interactions adhere strictly to physical laws.
    Collision handling, gravity, and object dynamics appear perfectly natural.
    \item \textbf{5:} Excellent. Minor, almost imperceptible physical inconsistencies
    (e.g., slight jitter on contact).
    \item \textbf{4:} Good. Largely realistic, but with noticeable violations
    (e.g., minor object penetration, slightly unnatural acceleration).
    \item \textbf{3:} Fair. Contains clear physical violations that are distracting
    (e.g., significant hand-through-object, unstable grasps).
    \item \textbf{2:} Poor. Multiple severe violations; the interaction often feels
    ``wrong'' (e.g., objects floating, unnatural collision responses).
    \item \textbf{1:} Non-physical. Completely breaks physical laws
    (e.g., objects teleporting, hands passing fully through solid objects).
\end{itemize}

\vspace{1ex}

\textbf{Key Evaluation Criteria:}
\begin{itemize}[itemsep=0.5ex]
    \item \textbf{Collision Handling:} Do objects and the agent interact without unnatural penetration?
    \item \textbf{Dynamics \& Gravity:} Do objects accelerate, decelerate, and fall realistically?
    \item \textbf{Contact \& Force:} Is the simulation of contact forces (e.g., friction, grasping) believable?
\end{itemize}

\vspace{1.5ex}
\textbf{Output Format (Strict JSON):}\\[0.5ex]
\texttt{\{"score": <integer\_score>, "rationale": "<brief\_justification>"\}}
\end{minipage}%
}
\end{minipage}
\hfill
\begin{minipage}{0.48\textwidth}
\centering
\fbox{%
\begin{minipage}{0.95\linewidth}
\small
\setlength{\parindent}{0pt}

\textbf{Box 2: $R_{\text{task}}$ (Task Completion) Evaluation Protocol}

\vspace{0.2cm}
\hrule
\vspace{0.2cm}

\textbf{System Prompt:}
You are an expert in logical reasoning and task planning.
Your task is to evaluate how successfully the agent completes the specified instruction.

\vspace{1.5ex}

\textbf{User Prompt:}\\[0.5ex]
\# Task Completion Evaluation\\
Task: \{task\_description\}

\vspace{1.5ex}

\textbf{Scoring Criteria (1--6 Scale):}
\begin{itemize}[itemsep=0.5ex]
    \item \textbf{6:} Perfect Success. The task is completed perfectly,
    efficiently, and exactly as instructed.
    \item \textbf{5:} Full Success. The task is fully completed, but with minor
    inefficiencies or slight deviations (e.g., suboptimal path).
    \item \textbf{4:} Partial Success. The main goal is achieved, but a key
    sub-task failed or was skipped.
    \item \textbf{3:} Attempted, but Failed. The agent attempts the correct task
    but fails to achieve the final goal (e.g., drops the object).
    \item \textbf{2:} Incorrect Task. The agent performs actions that are
    inconsistent with or irrelevant to the instruction.
    \item \textbf{1:} No Meaningful Action. The agent fails to initiate any
    logical action towards the goal.
\end{itemize}

\vspace{1ex}
\textbf{Key Evaluation Criteria:}
\begin{itemize}[itemsep=0.5ex]
    \item \textbf{Semantic Alignment:} Does the agent interact with the correct
    objects mentioned in the prompt?
    \item \textbf{Logical Sequence:} Are multi-step actions performed in the
    correct logical order?
    \item \textbf{Goal Achievement:} Is the final state of the environment
    consistent with the task goal?
\end{itemize}

\vspace{1.5ex}
\textbf{Output Format (Strict JSON):}\\[0.5ex]
\texttt{\{"score": <integer\_score>, "rationale": "<brief\_justification>"\}}
\end{minipage}%
}
\end{minipage}

\caption{VLM evaluation protocols for (left) physical realism and (right) task completion.}
\label{fig:vlm_protocols_12}
\end{figure*}

\begin{figure*}[t]
\centering
\begin{minipage}{0.48\textwidth}
\centering
\fbox{%
\begin{minipage}{0.95\linewidth}
\small
\setlength{\parindent}{0pt}

\textbf{Box 3: $R_{\text{embod}}$ (Embodiment Plausibility) Evaluation Protocol}

\vspace{0.2cm}
\hrule
\vspace{0.2cm}

\textbf{System Prompt:}
You are an expert in robot kinematics and biomechanics. Your task is to evaluate
the kinematic feasibility and naturalness of the agent's motion.

\vspace{1.5ex}

\textbf{User Prompt:}\\[0.5ex]
\# Embodiment Plausibility Evaluation\\
Task: \{task\_description\}

\vspace{1.5ex}

\textbf{Scoring Criteria (1--6 Scale):}
\begin{itemize}[itemsep=0.5ex]
    \item \textbf{6:} Perfect. The agent's motion is indistinguishable from a
    smooth, perfectly coordinated real-world robot.
    \item \textbf{5:} Natural. Motion is fluid and kinematically sound, with
    only trivial awkwardness.
    \item \textbf{4:} Acceptable. Motion is functional but exhibits noticeable
    non-human/non-robot-like artifacts (e.g., jitter, unnatural joint angles).
    \item \textbf{3:} Stiff/Robotic. Motion is rigid, lacks smoothness, and
    joints appear uncoordinated.
    \item \textbf{2:} Unnatural. Significant kinematic issues (e.g., impossible
    joint velocities, severe oscillation).
    \item \textbf{1:} Invalid. The agent's motion is broken, non-functional, or
    appears as a ``cloud of points''.
\end{itemize}

\vspace{1ex}

\textbf{Key Evaluation Criteria:}
\begin{itemize}[itemsep=0.5ex]
    \item \textbf{Kinematic Feasibility:} Are the joint angles, velocities, and
    accelerations plausible for the agent?
    \item \textbf{Smoothness \& Fluency:} Is the motion fluid, or is it jerky
    (high jerk)?
    \item \textbf{Coordination:} Do the different joints (e.g., arm, wrist,
    gripper) move in a coordinated fashion?
\end{itemize}

\vspace{1.5ex}

\textbf{Output Format (Strict JSON):}\\[0.5ex]
\texttt{\{"score": <integer\_score>, "rationale": "<brief\_justification>"\}}

\end{minipage}%
}
\end{minipage}
\hfill
\begin{minipage}{0.48\textwidth}
\centering
\fbox{%
\begin{minipage}{0.95\linewidth}
\small
\setlength{\parindent}{0pt}

\textbf{Box 4: $R_{\text{vis}}$ (Visual Quality) Evaluation Protocol}

\vspace{0.2cm}
\hrule
\vspace{0.2cm}

\textbf{System Prompt:}
You are an expert in video fidelity and aesthetics. Your task is to evaluate the
technical visual quality of the video, ignoring the agent's actions.

\vspace{1.5ex}

\textbf{User Prompt:}\\[0.5ex]
\# Visual Quality Evaluation\\
Task: \{task\_description\}

\vspace{1.5ex}

\textbf{Scoring Criteria (1--6 Scale):}
\begin{itemize}[itemsep=0.5ex]
    \item \textbf{6:} Excellent. Photorealistic, crisp, and perfectly stable.
    No visible artifacts.
    \item \textbf{5:} Good. High quality, but with minor issues
    (e.g., very slight blur, minimal noise).
    \item \textbf{4:} Acceptable. The video is clear, but suffers from
    noticeable issues (e.g., softness, minor flickering, shadows are off).
    \item \textbf{3:} Mediocre. Video is blurry, contains artifacts, or has
    unnatural lighting/textures.
    \item \textbf{2:} Poor. Significant visual artifacts, temporal
    inconsistency, or severe blur.
    \item \textbf{1:} Unwatchable. The video is corrupted, lacks coherence, or
    is dominated by visual noise.
\end{itemize}

\vspace{1ex}

\textbf{Key Evaluation Criteria:}
\begin{itemize}[itemsep=0.5ex]
    \item \textbf{Photorealism:} Do lighting, shadows, and textures appear realistic?
    \item \textbf{Artifact Suppression:} Is the video free of flickering,
    aliasing, or tearing?
    \item \textbf{Temporal Coherence:} Is the video stable and consistent from
    one frame to the next?
\end{itemize}

\vspace{1.5ex}

\textbf{Output Format (Strict JSON):}\\[0.5ex]
\texttt{\{"score": <integer\_score>, "rationale": "<brief\_justification>"\}}

\end{minipage}%
}
\end{minipage}

\caption{VLM evaluation protocols for (left) embodiment plausibility and (right) visual quality.}
\label{fig:vlm_protocols_34}
\end{figure*}

\subsection{Dimension Isolation Sampling Algorithm}
As stated in the main paper, random pairing of videos would introduce significant signal noise, as a video may excel in $R_{task}$ while failing in $R_{phys}$. This gradient interference would prevent our specialized HERO heads from learning their assigned functions.

To address this, we designed the Dimension Isolation Sampler, whose logic is detailed in the middle panel of \ref{fig:data_pipeline}. Grounded in the principle of mutual information minimization. The algorithm reframes pair selection as a constrained combinatorial optimization problem: find a pair $(v_A, v_B)$ that maximizes the variance in a single target dimension $k$ while minimizing the variance in all off-target dimensions $l \neq k$.Formally, given a target dimension $k \in \{phys, task, embod, vis\}$, we seek a pair $(v_A, v_B)$ such that:

\begin{enumerate}
\item Task Consistency: $\text{Prompt}(v_A) = \text{Prompt}(v_B)$
\item Target Variance (Condition 1 in \ref{fig:data_pipeline}): $|s_A^k - s_B^k| > \tau_{diff}$
\item Off-Target Covariance (Condition 2 in \ref{fig:data_pipeline}): $\forall l \neq k, |s_A^l - s_B^l| < \epsilon_{sim}$
\end{enumerate}
where $s_i^k$ is the VLM score for video $v_i$ in dimension $k$. Based on our dataset analysis, we set the hyperparameter $\tau_{diff} \ge 2.0$ and $\epsilon_{sim} \le 0.5$. The pseudocode for this sampling strategy is presented in Algorithm~\ref{alg:sampler}. This strategy provides the high-purity, decoupled training signals essential for the Dimensional Specificity Loss ($\mathcal{L}_{D}$) in Sec. 3.2.

\begin{algorithm}[t]
\caption{Dimension Isolation Sampling Algorithm}
\label{alg:sampler}
\small
\begin{algorithmic}[1]

\Require VLM-annotated video database
        $\mathcal{V} = \{ (v_i, \mathbf{s}_i, c_i) \}$
\Statex \hspace{\algorithmicindent}difference threshold
        $\tau_{\mathrm{diff}}$, similarity threshold
        $\epsilon_{\mathrm{sim}}$
\Ensure Dimension-isolated preference dataset
        $\mathcal{D}_{\mathrm{iso}}$

\State $\mathcal{D}_{\mathrm{iso}} \gets \emptyset$
\State Group $\mathcal{V}$ by task prompt $c$ into
       $\mathcal{G} = \{\mathcal{V}_{c_1}, \mathcal{V}_{c_2}, \dots\}$

\For{each task group $\mathcal{V}_c$ in $\mathcal{G}$}
  \For{each video $(v_A, \mathbf{s}_A, c)$ in $\mathcal{V}_c$}
    \For{each video $(v_B, \mathbf{s}_B, c)$ in $\mathcal{V}_c$}
      \If{$v_A = v_B$}
        \State \textbf{continue} \Comment{skip self-pairs}
      \EndIf
      \For{each dimension $k \in
            \{\text{phys}, \text{task}, \text{embod}, \text{vis}\}$}
        \State $\Delta_k \gets \lvert s_A^k - s_B^k \rvert$
        \State $\Delta_{\mathrm{other}} \gets
               \max_{l \neq k} \lvert s_A^l - s_B^l \rvert$
        \If{$\Delta_k > \tau_{\mathrm{diff}}$ \textbf{and}
            $\Delta_{\mathrm{other}} < \epsilon_{\mathrm{sim}}$}
          \If{$s_A^k > s_B^k$}
            \State $\mathcal{D}_{\mathrm{iso}} \gets
                   \mathcal{D}_{\mathrm{iso}} \cup \{(v_A, v_B, k)\}$
          \Else
            \State $\mathcal{D}_{\mathrm{iso}} \gets
                   \mathcal{D}_{\mathrm{iso}} \cup \{(v_B, v_A, k)\}$
          \EndIf
          \State \textbf{break} \Comment{use pair for a single dimension}
        \EndIf
      \EndFor
    \EndFor
  \EndFor
\EndFor

\Statex
\State \Comment{Additional strategies (e.g., Extreme Contrast,
        Nuance Pairs) further populate the dataset used to train
        $\mathcal{L}_{O}$.}
\State \Return $\mathcal{D}_{\mathrm{iso}}$

\end{algorithmic}
\end{algorithm}

\subsection{Final Dataset Statistics}
The combined output of our VLM annotation and sampling pipeline is a dataset of $\sim$235,000 preference pairs, derived from the $\sim$20K videos in the RH20T dataset. The composition of this dataset is detailed in Table~\ref{tab:dataset_stats}. This multi-faceted composition is essential: the dimension-isolated pairs provide the high-purity signal for $\mathcal{L}_{D}$, while the contrast and nuance pairs provide the calibration signal for $\mathcal{L}_{O}$.

\begin{table}[t]
\centering
\small
\caption{Composition of the 4D Embodied Preference Dataset.}
\label{tab:dataset_stats}
\resizebox{1\textwidth}{!}{
\begin{tabular}{p{0.36\linewidth} p{0.18\linewidth} p{0.38\linewidth}}
\toprule
\textbf{Preference Pair Type} &
\textbf{Approx.\ Quantity} &
\textbf{Purpose in HERO Training} \\
\midrule
Dimension Isolation ($R_{\text{phys}}$)  & $\sim\!55{,}000$ & Trains $R_{\text{phys}}$ head (via $\mathcal{L}_{D}$) \\
Dimension Isolation ($R_{\text{task}}$)  & $\sim\!60{,}000$ & Trains $R_{\text{task}}$ head (via $\mathcal{L}_{D}$) \\
Dimension Isolation ($R_{\text{embod}}$) & $\sim\!50{,}000$ & Trains $R_{\text{embod}}$ head (via $\mathcal{L}_{D}$) \\
Dimension Isolation ($R_{\text{vis}}$)   & $\sim\!30{,}000$ & Trains $R_{\text{vis}}$ head (via $\mathcal{L}_{D}$) \\
\midrule
Extreme Contrast Pairs         & $\sim\!20{,}000$ & Trains $\mathcal{L}_{O}$ (establishes score anchors) \\
Nuance Pairs (high similarity) & $\sim\!20{,}000$ & Trains $\mathcal{L}_{O}$ (improves fine-grained discrimination) \\
\midrule
\textbf{Total} & \textbf{$\sim 235{,}000$} & \\
\bottomrule
\end{tabular}
}
\vspace{-2pt}
\end{table}

\section{HERO Reward Model Architecture Details} 
\label{sec:supp_hero}
This section provides a detailed architectural breakdown of the HERO (HiErarchical Reward mOdel). The design of HERO is our solution to the ``\textit{Reward Barrier}'', and it is engineered to learn from the 4D Embodied Preference Dataset detailed in Section \ref{sec:supp_dataset}. We elaborate on the two core design principles introduced in Sec 3.2 of the main paper: (i) the hierarchical feature mapping strategy and (ii) the composite loss function that enables decoupled specialization and score calibration.
\subsection{Hierarchical Reward Awareness}
A foundational design principle of HERO is hierarchical reward awareness. We hypothesize that the four dimensions of embodied evaluation are not equally complex and require information from different levels of feature abstraction. For instance, evaluating $R_{phys}$ relies on low-level, fine-grained motion and geometry features. In contrast, evaluating $R_{task}$ requires high-level, abstract semantic understanding.

The InternVideo2-1B~\cite{wang2024internvideo2} backbone, with its progressive pre-training, produces a rich, hierarchical feature space. We exploit this by strategically mapping each specialized reward head to the feature block that is information-theoretically optimal for its designated task, rather than naively attaching all heads to the final output layer. This explicit mapping, detailed in Table~\ref{tab:feature_mapping}, ensures that each reward head receives the most salient feature representation for its evaluation task, which is critical for enabling functional specialization and minimizing gradient interference.

\begin{table}[t]
\centering
\small
\caption{Hierarchical Feature Mapping in HERO.}
\label{tab:feature_mapping}
\resizebox{0.9\textwidth}{!}{
\begin{tabular}{p{0.14\linewidth} p{0.18\linewidth} p{0.20\linewidth} p{0.10\linewidth} p{0.30\linewidth}}
\toprule
\textbf{Reward Head} &
\textbf{Eval.\ dim.} &
\textbf{IV2 block} &
\textbf{Feat.\ dim.} &
\textbf{Rationale} \\
\midrule
$R_{\text{phys}}$ &
Physical Realism &
Early-layer (Block 8) &
3200 &
Captures fine-grained motion, edges, and geometric cues essential for collision/penetration detection. \\
$R_{\text{embod}}$ &
Embodiment Plausibility &
Mid-layer (Block 16) &
3200 &
Encodes complex spatio-temporal motion patterns, ideal for assessing kinematic smoothness and coordination. \\
$R_{\text{task}}$ &
Task Completion &
Deep-layer (Block 32) &
1408 &
Represents high-level semantic concepts and logical state, necessary for understanding goal completion. \\
$R_{\text{vis}}$ &
Visual Quality &
Final pooled output &
768 &
Represents global aesthetics and frame-level coherence for overall fidelity assessment. \\
\bottomrule
\end{tabular}
}
\end{table}

\subsection{Reward Head and Critic Network}
The four HERO reward heads ($R_{phys}$, $R_{embod}$, $R_{task}$, $R_{vis}$) are implemented as lightweight Multi-Layer Perceptrons (MLPs) that project the high-dimensional features from their respective source layers (Table~\ref{tab:feature_mapping}) down to a single scalar reward value.

For completeness, we also detail the architecture of the VideoValueNetwork (Critic, $V_\psi$), which is a separate model trained from scratch during the HERO-FPO alignment phase. The $V_\psi$ model is a 3D-CNN designed to operate on raw video frames and estimate the expected trajectory reward, which is essential for computing the advantage $\hat{A}$ in Algorithm 1. The detailed architectures are presented in Table~\ref{tab:head_arch}.

\begin{table}[t]
\centering
\small
\caption{Network architectures for HERO reward heads and the FPO critic $V_\psi$.}
\label{tab:head_arch}
\renewcommand{\arraystretch}{1.05}  
\resizebox{0.9\textwidth}{!}{
\begin{tabular}{@{}p{0.18\linewidth} p{0.18\linewidth} p{0.58\linewidth}@{}}
\toprule
\textbf{Module} & \textbf{Input} & \textbf{Network architecture (PyTorch-style)} \\
\midrule
$R_{\text{phys}}$ head  & (B, 3200) &
Linear(3200, 512) $\rightarrow$ ReLU $\rightarrow$ Dropout(0.2) $\rightarrow$ Linear(512, 1) \\
$R_{\text{embod}}$ head & (B, 3200) &
Linear(3200, 512) $\rightarrow$ ReLU $\rightarrow$ Dropout(0.2) $\rightarrow$ Linear(512, 1) \\
$R_{\text{task}}$ head  & (B, 1408) &
Linear(1408, 512) $\rightarrow$ ReLU $\rightarrow$ Dropout(0.2) $\rightarrow$ Linear(512, 1) \\
$R_{\text{vis}}$ head   & (B, 768)  &
Linear(768, 256)  $\rightarrow$ ReLU $\rightarrow$ Dropout(0.2) $\rightarrow$ Linear(256, 1) \\
\midrule
$V_\psi$ critic & (B, 3, 16, 224, 224) &
Conv3d(3, 64, k=3, s=2) $\rightarrow$ ReLU $\rightarrow$ BatchNorm3d(64) \\
& & Conv3d(64, 128, k=3, s=2) $\rightarrow$ ReLU $\rightarrow$ BatchNorm3d(128) \\
& & Conv3d(128, 256, k=3, s=2) $\rightarrow$ ReLU $\rightarrow$ BatchNorm3d(256) \\
& & Conv3d(256, 256, k=3, s=2) $\rightarrow$ ReLU $\rightarrow$ BatchNorm3d(256) \\
& & AdaptiveAvgPool3d(1) $\rightarrow$ Flatten() $\rightarrow$ MLP(256, 128, 1) \\
\bottomrule
\end{tabular}
}
\end{table}

\subsection{Composite Loss Function Implementation}
As defined in the main paper, the total loss $\mathcal{L}_{HERO}$ is a composite objective designed to simultaneously solve two distinct challenges: functional specialization (via $\mathcal{L}_{D}$) and score calibration (via $\mathcal{L}_{O}$).
\begin{equation}
\mathcal{L}_{HERO} = \beta \cdot \mathcal{L}_{D} + (1 - \beta) \cdot \mathcal{L}_{O}
\label{eq:supp_loss_hero}
\end{equation}
In our final implementation, we set the specialization-calibration balance hyperparameter $\beta = 0.6$. This choice prioritizes (60\% weight) the Dimensional Specificity Loss ($\mathcal{L}_{D}$) to enforce the functional specialization of each head. The Overall Preference Regularizer ($\mathcal{L}_{O}$) acts as a 40\% calibration constraint to ensure all four head outputs remain on a comparable and meaningful numerical scale.

\subsubsection{Dimensional Specificity Loss}
$\mathcal{L}_{D}$ is the core objective that enforces functional specialization. Its efficacy relies on two key components derived from our 4D dataset, the dimensional mask ($\mathbf{M}_k$) and the adaptive weighting ($\mathbf{W}_k$).
\paragraph{Dimensional Mask ($\mathbf{M}_k$):}
This component functions as a binary gradient gate. For a given training pair $(v_A, v_B, k)$ tagged with target dimension $k$ (from Sec. \ref{sec:supp_dataset}), the mask $\mathbf{M}_k$ is $1$ if and only if the original VLM score difference $|s_A^k - s_B^k| > \tau_{margin}$.
\begin{itemize}
\item \textbf{Function:} This mask ensures that the specialized head $R_k$ only backpropagates loss on samples with a clear, unambiguous preference signal in its designated dimension, thereby preventing gradient interference from other dimensions.
\end{itemize}
\paragraph{Adaptive Weighting ($\mathbf{W}_k$):}
This component functions as an information gain prior, compelling the model to prioritize high-value training signals. It is computed as:
\begin{equation}
\mathbf{W}_k = (1 + w{iso}) \cdot |\Delta s_k|
\label{eq:supp_weight}
\end{equation}
The weight is composed of two factors:
\begin{itemize}
\item $|\Delta s_k| = |s_A^k - s_B^k|$: The absolute magnitude of the VLM score difference. This assigns a higher weight to high-confidence pairs where the VLM was more "certain" about the preference.
\item $w_{iso}$: The Dimension Isolation Boost Factor. If the sample was sourced from our high-purity "Dimension Isolation" set (Algorithm \ref{alg:sampler}), we set $w_{iso} = 1.0$; otherwise, $w_{iso} = 0$. This doubles the influence of our highest-quality, most informative samples, maximizing training efficiency.
\end{itemize}
\subsubsection{Overall Preference Regularizer}
Training with $\mathcal{L}_{D}$ alone ensures specialization but does not guarantee that the four heads output scores on a calibrated, comparable scale. For instance, $R_{phys}$ might learn to output in the range [0.1, 0.2] while $R_{task}$ outputs in [1.0, 5.0], rendering a simple sum of the heads meaningless.

$\mathcal{L}_{O}$ solves this score calibration problem. It is a standard Bradley-Terry loss applied to the final, combined scalar reward $R_{total} = \sum_{k} w_k R_k$. The weights $w_k$ used here are the same as those used during final inference (defined in Sec. \ref{sec:supp_benchmark_details}). This loss acts as a global regularizer, forcing all four heads to collaborate and project their specialized judgments onto a shared, meaningful numerical scale. The resulting frozen HERO model can then output a reliable $R_{total}$ signal for the HERO-FPO pipeline.

\section{HERO-FPO Algorithm and Theory} 
\label{sec:supp_fpo}

This section details the theoretical foundation and implementation specifics of our HERO-FPO (HERO-guided Flow Policy Optimization) pipeline. We first provide a detailed derivation for our core theoretical contribution, the CFM-Likelihood Proxy, which is foundational to solving the ``\textit{Algorithm Barrier}''. We then describe the critical engineering implementations required to integrate the multi-model training loop.
\subsection{The Intractability of Exact Likelihood}
Let $\pi_\theta(v|c)$ denote our flow-based video generation model, which transforms a simple prior distribution $p_0(z_0)$ to the complex video data distribution $p_1(v)$ via a time-dependent vector field $u_t(x)$. The exact log-likelihood of a generated video $v$ is given by the instantaneous change of variables formula:
\begin{equation}
    \log \pi_\theta(v|c) = \log p_0(z_0) - \int_0^1 \text{Tr}\left( \frac{\partial u_t^\theta(z_t, c)}{\partial z_t} \right) dt.
    \label{eq:intractable_ll}
\end{equation}
Computing the trace of the Jacobian $\text{Tr}(\nabla u_t^\theta)$ requires $\mathcal{O}(d^2)$ operations per ODE step. For high-resolution videos where $d \approx 10^6$, this integral is computationally prohibitive ($\mathcal{O}(d^2 \cdot T_{ODE})$), rendering standard PPO optimization infeasible.

\subsection{The CFM-ELBO Equivalence Theorem}
To bypass this bottleneck, we leverage the connection between Flow Matching and the Evidence Lower Bound (ELBO). The Conditional Flow Matching (CFM) objective used to train our model is defined as:
\begin{equation}
    \mathcal{L}_{CFM}(\theta) = \mathbb{E}_{t, q(v_t|v), p(v)}\left[ w(t) \| u_t^\theta(v_t, c) - u_t(v_t|v) \|^2 \right],
    \label{eq:cfm_loss}
\end{equation}
where $v_t$ is the noise-perturbed state at time $t$, and $u_t(v_t|v)$ is the conditional target vector field.

\vspace{1mm}
\noindent \textbf{Lemma 1 (CFM-ELBO Equivalence).} Following the derivation in \cite{mcallister2025flow}, for a diffusion noise schedule with uniform weighting $w(t)=1$, the weighted CFM loss is equivalent to the negative Evidence Lower Bound (ELBO) of the clean data plus a constant:
\begin{equation}
    \mathcal{L}_{CFM}(v; \theta, c) = - \text{ELBO}_\theta(v|c) + C(c),
    \label{eq:elbo_relation}
\end{equation}
where $C(c)$ is a constant that depends only on the noise schedule and data distribution, but is independent of the model parameters $\theta$.

\vspace{1mm}
\noindent \textbf{Proof Sketch.} The CFM objective can be viewed as a weighted denoising score matching objective. By reparameterizing the loss via Tweedie's formula, the objective decomposes into the variational lower bound of the log-likelihood plus a constant entropy term of the noise distribution. Since $\text{ELBO}_\theta(v|c) \le \log \pi_\theta(v|c)$, minimizing $\mathcal{L}_{CFM}$ maximizes a tight lower bound of the log-likelihood.

\subsection{Derivation of the HERO-FPO Update Rule}
Based on Lemma 1, we substitute the intractable log-likelihood with our tractable proxy. The standard PPO objective maximizes:
\begin{equation}
    J(\theta) = \mathbb{E}_{v \sim \pi_{\theta_{old}}}\left[ \frac{\pi_\theta(v|c)}{\pi_{\theta_{old}}(v|c)} \hat{A}(v, c) \right].
\end{equation}
Substituting $\log \pi_\theta(v|c) \approx \text{ELBO}_\theta(v|c) = -\mathcal{L}_{CFM}(v; \theta, c) + C(c)$, the importance sampling ratio $r(\theta)$ becomes:
\begin{align}
    r(\theta) &= \exp\left( \log \pi_\theta(v|c) - \log \pi_{\theta_{old}}(v|c) \right) \\
    &\approx \exp\left( \left[-\mathcal{L}_{CFM}^\theta + C(c)\right] - \left[-\mathcal{L}_{CFM}^{\theta_{old}} + C(c)\right] \right) \\
    &= \exp\left( \mathcal{L}_{CFM}(v; \theta_{old}, c) - \mathcal{L}_{CFM}(v; \theta, c) \right).
    \label{eq:final_ratio}
\end{align}
Crucially, the constant term $C(c)$ cancels out exactly. This proves that the difference in CFM losses provides a theoretically grounded estimate of the likelihood ratio, reducing the computational complexity from $\mathcal{O}(d^2 \cdot T)$ to $\mathcal{O}(d)$.

\subsection{Unbiased Gradient Direction Analysis}
A potential concern is that estimating $\mathcal{L}_{CFM}$ via Monte Carlo sampling (using $N$ noise samples) introduces a bias in the ratio $r(\theta)$ due to the convexity of the exponential function (Jensen's Inequality). 
However, as shown in \cite{mcallister2025flow}, the \textit{gradient direction} remains unbiased.

Let $\hat{r}(\theta)$ be the estimated ratio using Monte Carlo samples. The gradient of the PPO objective w.r.t $\theta$ is:
\begin{align}
    \nabla_\theta J &\propto \mathbb{E}\left[ \hat{A} \cdot \nabla_\theta \hat{r}(\theta) \right] \\
    &= \mathbb{E}\left[ \hat{A} \cdot \nabla_\theta \exp( \mathcal{L}_{old} - \mathcal{L}_{\theta} ) \right] \\
    &= \mathbb{E}\left[ \hat{A} \cdot \hat{r}(\theta) \cdot (- \nabla_\theta \mathcal{L}_{CFM}(v; \theta, c)) \right].
\end{align}
Since $\mathbb{E}[-\nabla_\theta \mathcal{L}_{CFM}] = \nabla_\theta \text{ELBO}$, the gradient update pushes the policy in the direction that maximizes the ELBO (and thus likelihood) of high-advantage samples. 
This theoretical property ensures that our HERO-FPO algorithm converges stably even when using a small number of noise samples ($N=5$) for efficiency, as empirically validated in our ablation studies.

\subsection{HERO-FPO Implementation Details}
Beyond the core theoretical proxy, several key engineering implementations are required to make the three-model (Actor, Critic, Reward) system tractable.
\subsubsection{Cosmos-SFT Domain Adaptation}
As stated in the main paper, we do not apply FPO directly to the base Cosmos-2B model. The base model is trained on general web videos, which have a significant domain gap with the contact-rich, kinematically-constrained physics of our RH20T robotics dataset.

To create a strong baseline and isolate the contribution of our RLHF pipeline, we first perform a supervised fine-tuning (SFT) step. We finetune the Cosmos-2B model on the Bridge V2 robotics dataset~\cite{walke2023bridgedata} using only the standard $L_{CFM}$ loss. This SFT phase adapts the model's prior to the robotics domain, resulting in our Cosmos-SFT baseline. The subsequent HERO-FPO alignment phase is therefore focused purely on optimizing for our 4D preferences, not on learning fundamental robotics priors from scratch.

\subsubsection{T5-11B Embedding Pre-computation}
The full HERO-FPO training loop—loading the Cosmos Actor, the HERO Reward Model, and the Critic—is extremely memory intensive. The largest component, however, is the T5-11B text encoder required by Cosmos for conditioning.

To make DDP training feasible on 8x A100 (80GB) GPUs, we implement a pre-computation strategy. We run the T5-11B model once offline to compute the text embeddings for all prompts in our RH20T training set. These embeddings are saved to disk. During FPO training, we replace the T5 model with a lightweight LRU (Least Recently Used) cache that loads these pre-computed embeddings on demand. This optimization completely removes the 44GB T5 model from the training loop, freeing up critical Video Random Access Memory (VRAM) for larger batch sizes and gradient accumulation.

\subsubsection{Cosmos-to-HERO Efficient GPU-based Video Adapter}
A critical engineering challenge is the format mismatch between the Actor (Cosmos) and the Reward Model (HERO), which operate at different resolutions and temporal rates.
\begin{itemize}
\item Cosmos Output ($\pi_\theta$): (B, 3, 16, 405, 720) (i.e., 16 frames at 720p, [-1, 1] range)
\item HERO Input ($R_{HERO}$): (B, 3, 8, 224, 224) (i.e., 8 frames at 224p, [0, 1] range)
\end{itemize}
We implement an efficient GPU-based VideoAdapter module that performs this conversion efficiently on the GPU before passing the video to the reward model:
\begin{enumerate}
\item Temporal Sampling: Sub-samples the 16-frame tensor to 8 frames using uniform indices.
\item Spatial Interpolation: Downscales the (405, 720) frames to (224, 224).
\item Value Normalization: Renormalizes the video tensor from Cosmos's [-1, 1] output range to HERO's [0, 1] input range via $v_{out} = (v_{in} + 1) / 2$.
\end{enumerate}

\section{ReWorldBench Evaluation Details} 
\label{sec:supp_benchmark_details}
This section provides a detailed exposition of ReWorldBench, the specialized benchmark introduced in the main paper. Standard video metrics (e.g., FVD, SSIM) are physics-agnostic and fundamentally unequipped to measure the critical failures within the ``Physics Uncanny Valley.'' ReWorldBench is therefore explicitly designed to probe and quantify model performance across the four key dimensions of embodied intelligence.
\subsection{Evaluation Dimensions and Task Design}
\label{ssec:eval_design}

The ReWorldBench evaluation task is defined as conditional video generation from an initial image and a text instruction (Image+Text-to-Video). Unlike standard video generation benchmarks that prioritize aesthetic adherence, ReWorldBench functions as a diagnostic suite designed to probe the model's ability to ground visual synthesis in physical reality. We evaluate generated videos along four orthogonal dimensions within our 4D preference space:

\begin{itemize}
    \item \textbf{Predictive Physical Reasoning ($S_{phys}$):} Assesses the model's fidelity to Newtonian mechanics. We specifically penalize \textit{dynamic} violations often ignored by standard metrics, such as object interpenetration (clipping), unnatural restitution (bounciness), spontaneous levitation, and inconsistencies in object permanence during occlusion.
    \item \textbf{Logical and Task Planning ($S_{task}$):} Evaluates semantic precision and sequential logic. Success is strictly binary based on goal achievement: did the agent interact with the \textit{correct} object (handling distractors) and execute sub-tasks in the \textit{correct temporal order}?
    \item \textbf{Kinematic Execution ($S_{embod}$):} Distinct from environmental physics, this dimension isolates the agent's own motion quality. We assess trajectories for robotic feasibility, penalizing kinematic hallucinations such as joint limit violations, end-effector teleportation, and high-frequency jitter (lack of smoothness).
    \item \textbf{Generative Fidelity ($S_{vis}$):} Evaluates foundational visual consistency. While less weighted in our embodied context, we assess photorealism and penalize generative artifacts like temporal flickering, texture warping, and background instability.
\end{itemize}

\vspace{1mm}
\noindent\textbf{Adversarial Prompt Engineering.}
Standard robotic datasets like RH20T~\cite{Fang2023arXiv} often contain simplistic instructions (e.g., ``Pick up the cup'') that mask model deficiencies. To expose these hidden failures, we implemented a systematic, two-stage Adversarial Expansion Pipeline using GPT-4o as an \textit{Adversarial Benchmark Architect}:

\begin{enumerate}
    \item \textbf{Ambiguity Injection (Curation):} We first curated implicit prompts (e.g., ``Tidy the desk'') that require the model to infer a multi-step plan from the visual state without explicit guidance. This tests the model's latent world knowledge.
    \item \textbf{Constraint Injection (Expansion):} For explicit tasks, we prompted GPT-4o to generate counterfactual and constrained variations. The meta-prompt instructed the LLM to: \textit{``Transform this simple instruction into a stress test that specifically challenges (a) collision avoidance, (b) negative constraints, or (c) fine-grained motor control.''}
\end{enumerate}

This process yielded a set of diagnostic probes targeting specific failure modes, as detailed below and exemplified in Table~\ref{tab:benchmark_prompts}:

\begin{itemize}
    \item \textbf{Probing $S_{phys}$ (Physics Stress Test):} We introduced interactions that require precise collision boundaries. For instance, transforming ``Move the block'' to ``Push the red block until it falls off the edge,'' forcing the model to simulate gravity and the transition from static to dynamic friction.
    \item \textbf{Probing $S_{task}$ (Logical Traps):} We injected semantic distractors and temporal dependencies. A simple ``Stack blocks'' became ``Pick up the blue cup, ignoring the red one,'' testing the model's ability to handle negative constraints—a common failure point for generative models.
    \item \textbf{Probing $S_{embod}$ (Kinematic Precision):} We added modifiers requiring fine motor control, such as ``Slowly and smoothly lift the fragile glass.'' This shifts evaluation from what is done to how it is done, penalizing the jerky, ballistic motions typical of diffusion hallucinations.
\end{itemize}

The resulting ReWorldBench comprises diverse, high-difficulty scenarios where visual plausibility alone is insufficient for high scores.

\begin{table}[h!]
\centering
\caption{\textbf{ReWorldBench Diagnostic Matrix.} Representative examples of how prompt design targets specific embodied failure modes.}
\label{tab:benchmark_prompts}
\small
\resizebox{\linewidth}{!}{
\begin{tabular}{l p{5.5cm} p{6.0cm}}
\toprule
\textbf{Dimension} & \textbf{Diagnostic Task Prompt} & \textbf{Targeted Failure Mode} \\ 
\midrule
$S_{phys}$ (Collision) & ``Grasp the red block and firmly push it \textbf{through} the blue block.'' & \textit{Adversarial Test}: Checks if the model hallucinates object penetration instead of simulating physical resistance/blocking. \\ 
$S_{phys}$ (Gravity) & ``Pick up the apple, lift it high, and then \textbf{release} it in mid-air.'' & \textit{Dynamics Test}: Checks for "floating object" hallucinations vs. correct gravitational acceleration. \\ 
$S_{task}$ (Logic) & ``\textbf{First}, pick up the sponge, \textbf{then} wipe the counter, and \textbf{finally} place it in the sink.'' & \textit{Memory Test}: Probes long-horizon temporal consistency and instruction following. \\ 
$S_{task}$ (Semantics) & ``There are two cups. Pick up the blue cup, \textbf{ignoring} the red one.'' & \textit{Negation Test}: Evaluates fine-grained concept binding and resistance to semantic distractors. \\ 
$S_{embod}$ (Kinematics) & ``\textbf{Slowly and smoothly} reach for the fragile wine glass.'' & \textit{Control Test}: Penalizes high-jerk, unnatural "teleporting" motions common in video generation. \\ 
$S_{embod}$ (Grasping) & ``Use a \textbf{pincer grasp} to pick up the small screw.'' & \textit{Fine-Motor Test}: Checks for plausible hand-object contact points vs. generic "magnetic" grasping. \\
\bottomrule
\end{tabular}
}
\end{table}

\subsection{VLM-as-Judge Evaluation Protocol} 
\label{ssec:vlm_judge}
To ensure a reproducible, interpretable, and rigorous evaluation, we employ a ``VLM-as-Judge'' protocol. While human evaluation remains the gold standard, it is not scalable for large-scale benchmarking. We utilize GPT-4o as our expert proxy, chosen for its state-of-the-art spatio-temporal reasoning capabilities. 
However, standard VLM scoring suffers from inherent biases and hallucinations. To mitigate these, we designed a strict protocol centered on rationale-first reasoning and criteria-guided scoring.

\vspace{1mm}
\noindent\textbf{The Rationale-First Mechanism.} 
A core innovation of our protocol is the mandatory \textit{``Rationale-First''} constraint. As shown in Box 5, we explicitly forbid the model from outputting a score immediately. Instead, the model must first generate a detailed, step-by-step textual analysis of the video. 
For example, in the $S_{phys}$ dimension, the model is prompted to \textit{``Analyze collisions, gravity, and object interactions''} before deciding on a score. This forces the model to ground its judgment in observable video evidence, significantly reducing ungrounded hallucinations and improving correlation with human judgment.

\vspace{1mm}
\noindent\textbf{Fine-Grained 10-Point Scale.} 
Unlike the 6-point Likert scale used for our training dataset (which prioritizes robust preference learning), ReWorldBench employs a finer 1-10 scale to maximize discriminability between competitive SOTA models.
\begin{itemize}
    \item 1-3 (Failure): Critical physical violations (e.g., object teleportation) or complete task failure.
    \item 4-6 (Mediocre): Visually coherent but physically flawed (e.g., sliding without friction, minor interpenetration).
    \item 7-8 (Good): Correct physics and logic, with minor visual artifacts.
    \item 9-10 (Excellent): Indistinguishable from a real-world robotic execution.
\end{itemize}

\vspace{1mm}
\noindent\textbf{Role-Playing \& Structured Output.} 
To further align the VLM's focus, we utilize System Prompt Engineering to inject specific personas (e.g., \textit{``You are an expert in Newtonian mechanics...''}). Finally, to ensure automated processing, the output is strictly constrained to a JSON schema containing separate fields for the rationale string and the score integer. The full prompt structure is detailed in Box 5.

\begin{figure*}[t]
\centering
\fbox{
\begin{minipage}{0.95\textwidth}
\textbf{Box 5: ReWorldBench Rigorous Evaluation Protocol (Expanded)}
\vspace{0.2cm}
\hrule
\vspace{0.2cm}
\small

\noindent \textbf{System Prompt:} 
You are a strictly objective, highly critical expert evaluator for embodied AI robotics. Your goal is to identify subtle failures in generated videos that indicate a lack of physical grounding. You must penalize hallucinations, physical violations, and logical errors severely. \textbf{Avoid Sycophancy Bias:} do not give high scores to be polite. A score of 10 requires perfection indistinguishable from a real-world recording.

\vspace{0.2cm}
\noindent \textbf{User Prompt:}\\
\textbf{\# Embodied Video Evaluation}\\
Instruction: \{task\_description\}

\vspace{0.2cm}
\noindent \textbf{\#\# Scoring Rubric (Reference):}\\
\begin{itemize}[leftmargin=*]
    \item \textbf{1-3 (Failure):} Critical violation (e.g., object passes through table, wrong object grasped, gravity ignored).
    \item \textbf{4-6 (Mediocre):} Task attempted but physics is "slushy" (objects slide without friction) or motion is glitchy.
    \item \textbf{7-8 (Good):} Correct logic and physics. Minor visual artifacts (blur, background flickering) allowed.
    \item \textbf{9-10 (Perfect):} Flawless execution. Physics, lighting, and kinematics are indistinguishable from reality.
\end{itemize}

\vspace{0.2cm}
\noindent \textbf{\#\# Evaluation Protocol (Rationale-First):}\\
Step 1: Analyze the video frame-by-frame and write a critical evidence-grounded rationale for each dimension.
Step 2: Assign a score (1-10) based on the Rubric.

\vspace{0.1cm}
\noindent --- RATIONALE GUIDELINES ---
\vspace{0.1cm}

\noindent \textbf{1. Physical Realism ($S_{phys}$):}
Focus on \textbf{Dynamics}. Do objects vanish (Object Permanence)? Do they pass through each other (Interpenetration/Clipping)? Does the object float or slide unnaturally (Gravity/Friction)?

\vspace{0.1cm}
\noindent \textbf{2. Task Completion ($S_{task}$):}
Focus on \textbf{Logic}. Did the agent pick the \textit{correct} object and ignore distractors (Negative Constraints)? Was the sequence of sub-tasks correct (Temporal Order)?

\vspace{0.1cm}
\noindent \textbf{3. Embodiment Plausibility ($S_{embod}$):}
Focus on \textbf{Kinematics}. Is the robot arm's motion continuous? Check for "teleporting" joints or high-frequency jitter. Is the grasp physically valid (surface contact) or just a "magnetic" attachment?

\vspace{0.1cm}
\noindent \textbf{4. Visual Quality ($S_{vis}$):}
Focus on \textbf{Fidelity}. Is the video coherent? Are there severe artifacts like tearing or warping?

\vspace{0.1cm}
\noindent --- OUTPUT FORMAT (JSON ONLY) ---\\
\texttt{\{ \\
  \phantom{..}"rationale\_phys": "...", \\
  \phantom{..}"score\_phys": <int>, \\
  \phantom{..}"rationale\_task": "...", \\
  \phantom{..}"score\_task": <int>, \\
  \phantom{..}"rationale\_embod": "...", \\
  \phantom{..}"score\_embod": <int>, \\
  \phantom{..}"rationale\_vis": "...", \\
  \phantom{..}"score\_vis": <int> \\
\}}

\end{minipage}
}
\end{figure*}

\subsection{Final Benchmark Score $S_{ReWorld\text{-}Bench}$ Rationale}
To provide a single, comprehensive metric for model comparison, we compute the final $S_{ReWorld\text{-}Bench}$ score. Each dimension is first scored on a 1--10 scale; we then report the weighted aggregate on a 0--100 scale for readability:
\begin{equation}
S_{ReWorld\text{-}Bench} = 10 \times (0.4 \times S_{task} + 0.3 \times S_{embod} + 0.2 \times S_{phys} + 0.1 \times S_{vis}).
\label{eq:supp_benchmark_score}
\end{equation}
This weighting scheme is a deliberate design choice reflecting the priorities of embodied intelligence over standard video generation. The rationale is as follows:
\begin{itemize}
\item \textbf{$S_{task}$ (40\%):} Task Completion receives the highest weight, as goal-directedness is the primary objective of an embodied agent. A physically realistic action that fails the task is ultimately unsuccessful.
\item \textbf{$S_{embod}$ (30\%):} Embodiment Plausibility is weighted heavily. How an agent moves is critical for sim-to-real transfer and grounding. A kinematically impossible motion, even if it achieves the goal, is a failure of the video generation model.
\item \textbf{$S_{phys}$ (20\%):} Physical Realism is a core constraint. A ``trustworthy'' EWM must adhere to the laws of physics, forming the foundation upon which tasks can be learned.
\item \textbf{$S_{vis}$ (10\%):} Visual Quality receives the lowest weight. While a prerequisite for grounding, modern generative models already excel here. Our benchmark prioritizes the (often-neglected) functional dimensions over pure aesthetics.
\end{itemize}

\subsection{Human Evaluation Protocol}
To validate the reliability of our automated metrics and confirm the perceptual superiority of ReWorld, we conducted a rigorous human evaluation campaign. Unlike crowdsourcing platforms which often yield high-variance results for technical physics tasks, we employed an Expert Review protocol.

\noindent\textbf{Evaluator Selection.} We recruited 5 domain experts to serve as annotators. All participants are Ph.D. candidates or researchers specializing in robotics, computer vision, or physics simulation. This ensures that the evaluators possess the necessary domain knowledge to identify subtle physical violations that laypersons might overlook.

\noindent\textbf{Blind A/B Testing Protocol.}
We established a rigorous double-blind side-by-side comparison protocol. To eliminate potential biases, the evaluation process adhered to the following strict standards:
\begin{itemize}
\item Anonymity: All video files were anonymized. Neither the model names nor the generation methods were visible to the evaluators during the assessment.
\item Randomization: To prevent position bias, the placement (left/right) of the ReWorld rollout and the baseline rollout was randomly shuffled for each query.
\item Fine-Grained Inspection: Evaluators were instructed to examine videos with frame-level control, ensuring critical physical details were thoroughly analyzed.
\end{itemize}

\noindent\textbf{Scoring Criteria.} Evaluators were presented with the same text prompt used for generation and asked to select the better video based on the 4D criteria defined in the main paper. For each pair, they could choose: \textit{``Left is Better''}, \textit{``Right is Better''}, or \textit{``Tie''}. To align human judgment with our HERO model, experts were explicitly instructed to prioritize: 
\begin{enumerate} 
\item Physics (Primary): Any object penetration or gravity violation results in an immediate loss, regardless of visual quality. 
\item Task (Primary): Failure to complete the instruction results in a loss. 
\item Visuals (Secondary): Visual artifacts are penalized only if physical/task logic is comparable. 
\end{enumerate} 
This priority hierarchy ensures that the human evaluation targets the ``Physics Uncanny Valley'' rather than superficial aesthetic appeal.

\noindent\textbf{Participant Instructions and Compensation.}
Evaluators were instructed that the study was a technical quality assessment of generated robot videos. For each anonymized pair, they reviewed the shared prompt, inspected both videos at their own pace, and selected one of three responses: \textit{Left is Better}, \textit{Right is Better}, or \textit{Tie}. The evaluation did not collect personal, medical, behavioral, or sensitive participant data. Compensation followed local institutional practice for expert technical evaluation; no open crowdsourcing marketplace was used.

\section{Implementation and Hyperparameter Details} 
\label{sec:supp_hyper}
We provide the complete implementation details, training configurations, and hyperparameters for all stages of the ReWorld framework to ensure full reproducibility.
\subsection{HERO Training Hyperparameters}
The configuration for training our HERO reward model is detailed in Table~\ref{tab:hyper_hero}. The model was trained on our 4D Embodied Preference Dataset ($\sim$235K pairs), as described in Section \ref{sec:supp_dataset}.

\begin{table}[t]
\centering
\caption{Hyperparameters for HERO Reward Model Training.}
\label{tab:hyper_hero}

\begin{tabularx}{\linewidth}{l X} 
\toprule
\textbf{Parameter} & \textbf{Value} \\
\midrule
\multicolumn{2}{l}{\textbf{Model Architecture}} \\
Video Backbone & InternVideo2-1B \\
Backbone Parameters & Frozen \\
Head Architecture & See Table~\ref{tab:head_arch} \\
Head Dropout & 0.2 \\
\midrule
\multicolumn{2}{l}{\textbf{Training \& Optimization}} \\
Optimizer & AdamW \\
Learning Rate & 1e-5 \\
Weight Decay & 0.01 \\
Adam Betas & (0.9, 0.999) \\
Total Batch Size & 64 \\
Batch Size per GPU & 8 (on 8 GPUs) \\
Max Epochs & 100 \\
LR Scheduler & Cosine Annealing w/ 1-epoch warmup \\
Gradient Clip Norm & 1.0 \\
Mixed Precision & bfloat16 \\
\midrule
\multicolumn{2}{l}{\textbf{Composite Loss Function}} \\
$\beta$ (Loss Weight) & 0.6 (60\% for $\mathcal{L}_{D}$) \\
$1 - \beta$ (Weight) & 0.4 (40\% for $\mathcal{L}_{O}$) \\
$\tau_{margin}$ (Thres.) & 0.2 \\
$w_{iso}$ (Iso. Boost) & 1.0 (2x for isolated) \\
BT Temp. ($\tau$) & 0.5 \\
\bottomrule
\end{tabularx}
\end{table}

\subsection{HERO-FPO Training Hyperparameters} 
The configuration for the HERO-FPO alignment phase is detailed in Table~\ref{tab:hyper_fpo}. This phase optimizes the Cosmos-SFT actor policy using the frozen HERO reward model.

\begin{table}[t]
\centering
\caption{Hyperparameters for HERO-FPO Policy Alignment Training.}
\label{tab:hyper_fpo}
\begin{tabularx}{\linewidth}{l X}
\toprule
\textbf{Parameter} & \textbf{Value} \\
\midrule
\multicolumn{2}{l}{\textbf{Models \& Hardware}} \\ 
Actor ($\pi_\theta$) & Cosmos-2B (SFT on Bridge V2) \\ 
Reward ($R_{HERO}$) & HERO (InternVideo2-1B, Frozen) \\
Critic ($V_\psi$) & VideoValueNetwork (See Table~\ref{tab:head_arch}) \\
Hardware & 8 $\times$ NVIDIA A100 (80GB) \\
\midrule
\multicolumn{2}{l}{\textbf{Training \& Optimization}} \\ 
Total Epochs (N) & 100 \\
PPO Epochs / Iter. (K) & 4 \\
Batch Size per GPU & 2 \\
Grad. Accum. Steps & 8 \\
Effective Batch Size & 128 (2 $\times$ 8 GPUs $\times$ 8 steps) \\
Optimizer & AdamW \\
Actor LR ($\pi_\theta$) & 1e-5 (Cosmos DiT) \\
Critic LR ($V_\psi$) & 3e-4 (Value Network) \\
LR Scheduler & Cosine Annealing \\
Warmup Steps & 1000 \\
Weight Decay & 1e-4 \\
Mixed Precision & bfloat16 \\
\midrule
\multicolumn{2}{l}{\textbf{PPO Algorithm}} \\
Clip Epsilon ($\epsilon$) & 0.2 \\
GAE Gamma ($\gamma$) & 0.99 \\
GAE Lambda ($\lambda$) & 0.95 \\
Value Loss Coeff. ($c_v$) & 0.5 \\
Entropy Coeff. & 0.01 \\
Max Gradient Norm & 1.0 \\
Policy Sync Interval & 1 Iteration ($\pi_{\theta_{old}} \gets \pi_\theta$) \\
\midrule
\multicolumn{2}{l}{\textbf{CFM-Likelihood Proxy}} \\
CFM Sampler ($N$) & 5 (validated in Table 4) \\
EDM Weighting & True \\
$\sigma_{data}$ (Data Scale) & 1.0 \\
\bottomrule
\end{tabularx}
\end{table}

\subsection{Infrastructure and Training Runtimes}
All experiments were conducted on a single server node equipped with 8 NVIDIA A100 (80GB) GPUs, interconnected via NVLink. The approximate runtimes for each stage of the ReWorld pipeline are as follows:
\begin{itemize}
\item Stage 1: Generating the 4D annotations for $\sim$20K videos using the GPT-4o API and subsequently sampling the $\sim$235K preference pairs took approximately 48 hours.
\item Stage 2: Training the HERO reward model on the $\sim$235K dataset for 20 epochs took approximately 24 hours.
\item Stage 3: The initial Supervised Fine-Tuning (SFT) of Cosmos-2B on the Bridge V2 dataset took approximately 50 hours. The final HERO-FPO alignment phase, run for 100 epochs, took approximately 72 hours.
\end{itemize}

\section{Limitations, Broader Impacts, and Asset Documentation}
\label{sec:supp_responsible}

\subsection{Limitations}
ReWorld is designed for embodied video generation and is primarily validated on robotic manipulation scenarios from RH20T and Bridge V2. Although the 4D reward decomposition is general in principle, performance in other embodied domains such as navigation, deformable-object manipulation, or human-centered interaction may require additional preference data, prompts, and benchmark coverage. The method also relies on VLM-based annotation and VLM-as-judge evaluation, which can inherit model biases, prompt sensitivity, and failure modes in fine-grained temporal reasoning. We mitigate this with dimension-specific rubrics, strict JSON schemas, rationale-first evaluation, and expert validation, but VLM judgments should not be treated as a perfect substitute for large-scale human evaluation.

HERO-FPO uses the CFM-likelihood proxy rather than exact flow-model likelihood. This proxy makes PPO-style alignment computationally feasible for high-dimensional video, but it remains an approximation and may introduce bias when the CFM objective is a loose likelihood surrogate or when reward optimization moves far from the SFT policy distribution. The framework is also computationally intensive: annotation, reward training, Cosmos SFT, and HERO-FPO require multi-GPU resources. Finally, any learned reward model can be vulnerable to reward hacking; future work should study stronger out-of-distribution diagnostics, adversarial benchmark construction, and more sample-efficient alignment.

\subsection{Broader Impacts}
The positive goal of ReWorld is to improve the physical reliability and task consistency of embodied world models, which may benefit robotic simulation, policy pretraining, diagnostic benchmarking, and safer offline evaluation before deploying agents in real environments. A more reliable embodied video generator can help expose physical and logical failures that are not captured by standard visual metrics.

The same capability also carries risks. Increasingly realistic generated robot videos may be misinterpreted as evidence of real-world robot competence, and over-reliance on generated rollouts could lead to unsafe conclusions if physical or causal errors remain. VLM-based annotation may encode systematic biases from the underlying model or prompt design. To reduce these risks, we emphasize diagnostic evaluation, disclose the use of VLMs, avoid public release of unrestricted high-risk checkpoints or generated-video datasets at submission time, and require any future release to follow the licenses and usage restrictions of the underlying assets.

\subsection{Existing and Derived Assets}
ReWorld uses existing datasets, models, and services for academic research. RH20T is used as the primary embodied video source for annotation and evaluation; Bridge V2 is used for Cosmos domain-adaptation SFT; Cosmos-2B is the flow-based video generator backbone; InternVideo2 is used as the HERO video backbone; CogVideoX and Wan2.1 are used as external baselines; GPT-4o is used for VLM-based annotation, benchmark prompt expansion, and VLM-as-judge evaluation; standard video-evaluation components are used for FVD, SSIM, DINO Similarity, PSNR, and DreamSim.

The derived assets introduced by this work include the 4D embodied preference dataset, ReWorldBench prompts and evaluation annotations, trained HERO reward-model artifacts, HERO-FPO aligned model outputs, and generated videos used in evaluation. These derived assets are documented through the construction protocols, prompt templates, scoring rubrics, model architectures, hyperparameters, and evaluation procedures in the paper and appendix. They are not publicly released at submission time. Any future release of code, checkpoints, preference annotations, or benchmark assets will respect the licenses, access constraints, and terms of use of the underlying datasets, pretrained models, software, and VLM services.

\section{Additional Qualitative Results and Analysis}
\label{sec:supp_qual}

\begin{figure*}[t]
  \centering
  \includegraphics[width=1.0\linewidth]{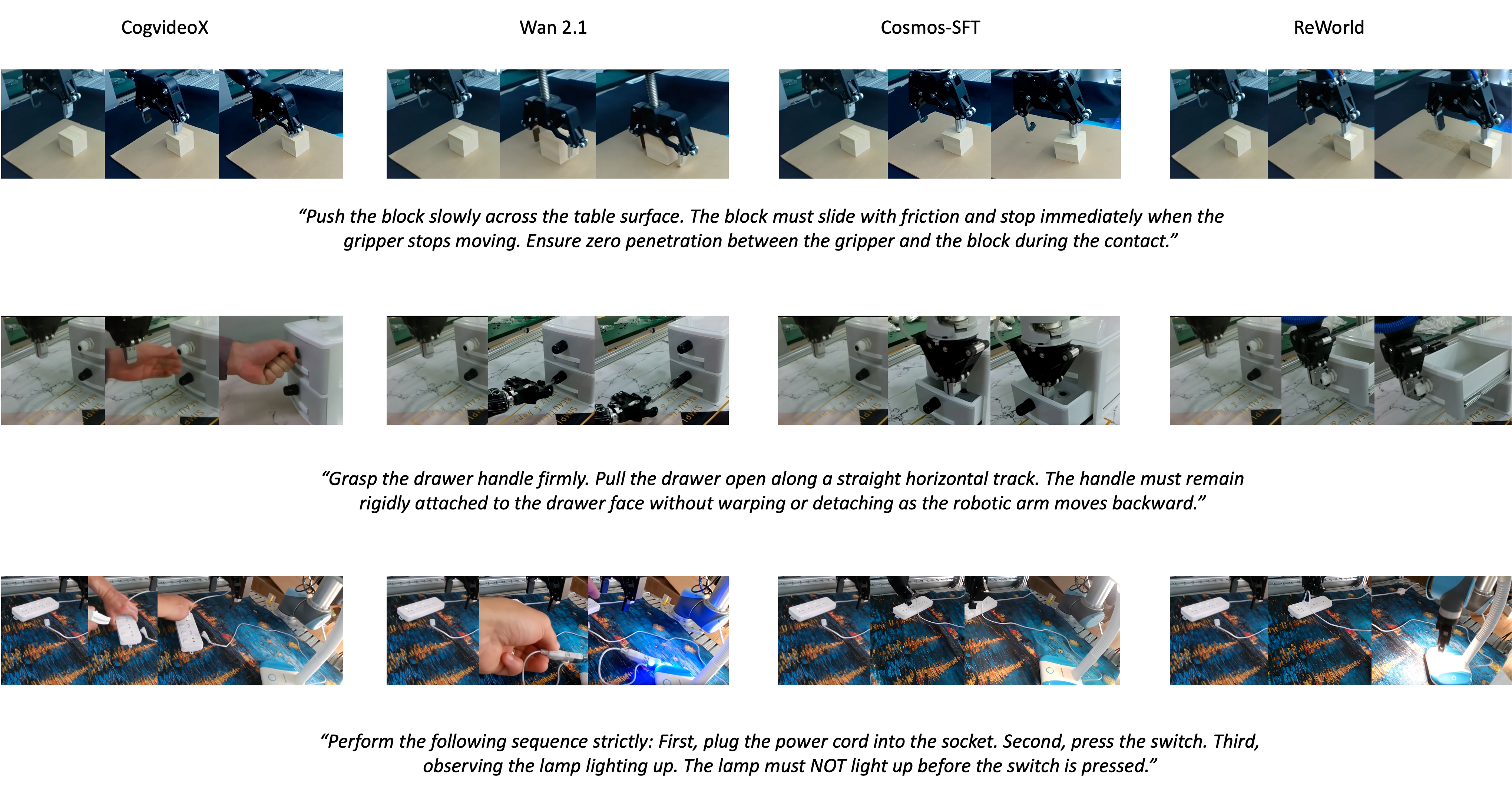}
  \vspace{-3mm}
  \caption{\textbf{Qualitative comparison against SOTA baselines on the ReWorldBench diagnostic suite.} 
  We visualize generated rollouts for three distinct task primitives: 
  \textbf{(Top Row) Rigid Body Interaction:} Testing contact physics and friction using a ``Pushing'' task. Baselines often exhibit ``ghost touches'' or sliding without friction.
  \textbf{(Middle Row) Articulated Object Manipulation:} Testing kinematic constraints using a ``Drawer Opening'' task. Baselines struggle to maintain the rigid connection between the handle and the drawer.
  \textbf{(Bottom Row) Sequential Logic:} Testing long-horizon causality using a ``Plug \& Switch'' task. Baselines fail the negative constraint, often lighting up the lamp before the switch is pressed.
  \textbf{ReWorld (Rightmost Column)} consistently adheres to physical, kinematic, and logical constraints specified in the diagnostic prompts.}
  \label{fig:sota_comparison}
\end{figure*}

This section provides granular visual evidence to substantiate the quantitative gains reported in the main paper. 
Crucially, all visualizations presented in ~\ref{fig:sota_comparison} are sampled directly from the ReWorldBench diagnostic suite.
Unlike standard prompts, these adversarial prompts impose strict physical and logical constraints to stress-test model capabilities. Below, we detail the specific prompts used and analyze the behaviors of ReWorld versus SOTA baselines, including Cosmos-SFT, CogVideoX~\cite{yang2025cogvideoxtexttovideodiffusionmodels}, and Wan2.1~\cite{wan2025wanopenadvancedlargescale}.

To ensure a rigorous and fair comparison, all baselines are evaluated under a consistent Image-to-Video (I2V) setting, conditioned on the initial frame and text instruction from the ReWorldBench test set. We establish Cosmos-SFT as our primary internal baseline by fine-tuning the official Cosmos-2B model~\cite{alhaija2025cosmos} on the Bridge V2 dataset~\cite{walke2023bridgedata} using standard CFM loss; this explicitly isolates the contribution of our HERO-FPO alignment by sharing the exact backbone and data distribution. For external state-of-the-art comparisons, we adopt CogVideoX-5B~\cite{yang2025cogvideoxtexttovideodiffusionmodels} and Wan2.1-14B~\cite{wan2025wanopenadvancedlargescale} using their official repositories and inference pipelines, adapting input frames to their native resolution requirements. While these models represent the pinnacle of open-domain generation, they are evaluated here in a zero-shot manner to underscore the necessity of domain-specific physical alignment. Finally, for consistent metric computation, all generated outputs are spatially resized and temporally downsampled to the canonical $224 \times 224 \times 16$ format.

\begin{figure*}[t]
  \centering
  \includegraphics[width=0.8\linewidth]{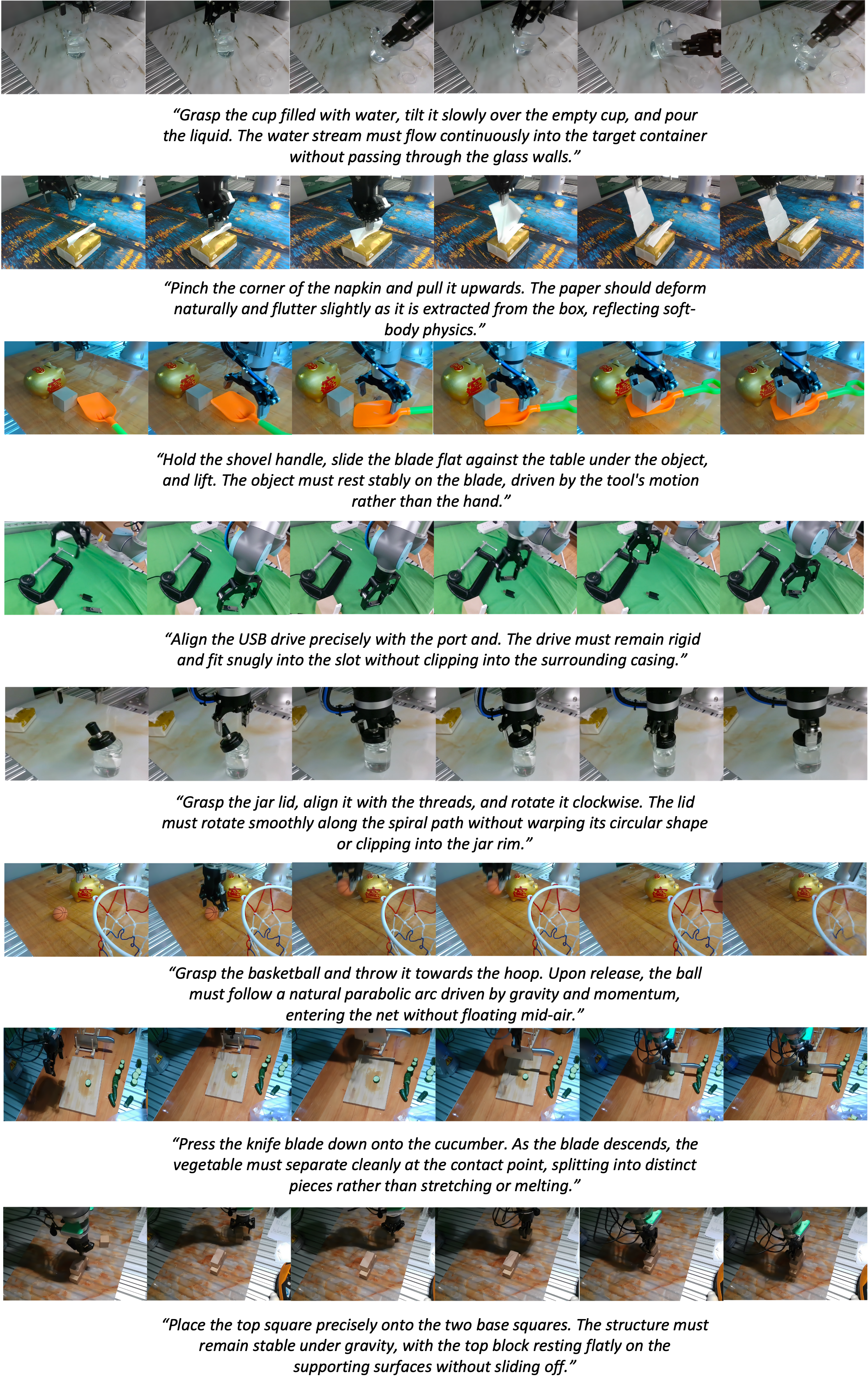}
  \caption{\textbf{Gallery of diverse embodied skills generated by ReWorld.} 
The figure showcases the model's generalization across eight distinct physical domains, ranging from \textbf{complex material properties} (fluids, soft bodies) to \textbf{dynamic laws} (rotation, projectile motion) and \textbf{geometric constraints} (topology change, high-precision assembly). 
Detailed diagnostic prompts are provided below each sequence.}
  \label{fig:gallery}
\end{figure*}

\noindent\textbf{Case I: Physics Stress Test (Contact \& Friction)}

Source Task: RH20T Task-0006.
Diagnostic Prompt: \textit{``Push the block slowly across the table surface. The block must slide with friction and stop immediately when the gripper stops moving. Ensure zero penetration between the gripper and the block during the contact.''}

As shown in the Top Row of \ref{fig:sota_comparison}:
\begin{itemize}
    \item \textbf{Baselines Failure:} Cosmos-SFT and CogVideoX exhibit the ``Ghost Touch'' phenomenon. The block begins to slide before the gripper makes physical contact, or continues to drift as if on ice after the gripper stops. This indicates a failure to model friction and contact causality.
    \item \textbf{ReWorld Success:} ReWorld precisely synchronizes the gripper's contact frame with the object's motion initiation. The block stops naturally due to friction, validating the efficacy of the $R_{phys}$ head in capturing Newtonian dynamics.
\end{itemize}

\noindent\textbf{Case II: Kinematic Integrity Test (Rigid Bodies)}

Source Task: RH20T Task-0029.
Diagnostic Prompt: \textit{``Grasp the drawer handle firmly. Pull the drawer open along a straight horizontal track. The handle must remain rigidly attached to the drawer face without warping or detaching as the robotic arm moves backward.''}

The Middle Row of \ref{fig:sota_comparison} visualizes the kinematic integrity of opening a drawer:
\begin{itemize}
    \item \textbf{Baselines Failure:} While Wan 2.1 generates high-fidelity textures, it suffers from geometric hallucinations. As the arm pulls back, the handle visually warps or momentarily detaches from the drawer face, violating the rigid body constraint of the articulated object.
    \item \textbf{ReWorld Success:} Our model generates a physically valid trajectory. The robotic arm, handle, and drawer move as a coherent kinematic chain along the prismatic joint, satisfying the \textit{Embodiment Plausibility} ($R_{embod}$) criteria.
\end{itemize}

\noindent\textbf{Case III: Causal Logic Trap (Sequential Planning)}

Source Task: RH20T Task-0127.
Diagnostic Prompt: \textit{``Perform the following sequence strictly: First, plug the power cord into the socket. Second, press the switch. Third, observing the lamp lighting up. The lamp must NOT light up before the switch is pressed.''}

The Bottom Row of \ref{fig:sota_comparison} tests long-horizon causal reasoning:
\begin{itemize}
    \item \textbf{Baselines Failure:} Most models fail the ``Negative Constraint''. For instance, CogVideoX and Cosmos-SFT tend to hallucinate the lamp turning on as soon as the hand approaches the plug, violating the causal chain of the electric circuit.
    \item \textbf{ReWorld Success:} ReWorld correctly models the hidden state of the system. The lamp remains off during the plugging phase and illuminates strictly after the switch interaction is completed, demonstrating superior Task Completion ($R_{task}$) guided by deep semantic features.
\end{itemize}

\section{Gallery of Diverse Embodied Skills}
\label{sec:gallery}

To demonstrate the breadth and generalization of ReWorld, \ref{fig:gallery} presents an uncurated gallery of generated rollouts across eight distinct physical domains. 
Moving beyond standard rigid-body interactions, the model successfully simulates complex material properties, including fluid viscosity (pouring water) and soft-body deformation (pulling napkins). 
Furthermore, ReWorld strictly adheres to Newtonian dynamics—correctly modeling momentum in projectile motion and rotational constraints in screwing tasks—while maintaining geometric precision during topological changes (cutting) and static equilibrium (stacking). 
These results validate that ReWorld acquires a generalized physics prior capable of handling a wide spectrum of embodied interactions.


\end{document}